\documentclass[lettersize,journal]{IEEEtran}
\usepackage{amsmath,amsfonts}
\usepackage{algorithmic}
\usepackage{algorithm}
\usepackage{array}
\usepackage[caption=false,font=normalsize,labelfont=sf,textfont=sf]{subfig}
\usepackage{textcomp}
\usepackage{stfloats}
\usepackage{url}
\usepackage{verbatim}
\usepackage{graphicx,subfig}
\usepackage{cite}
\usepackage{bbm}
\usepackage{booktabs}

\usepackage{tikz}
\usetikzlibrary{calc}
\usepackage{subcaption}
\usepackage[justification=centering]{caption}

\usepackage{hyperref}

\begin{document}

\title{CDFL: Efficient Federated Human Activity Recognition using Contrastive Learning and Deep Clustering}

%\author{Ensieh Khazaei,~\IEEEmembership{Staff,~IEEE,}
%\author{IEEE Publication Technology,~\IEEEmembership{Staff,~IEEE,}
\author{Ensieh Khazaei, Alireza Esmaeilzehi,~\IEEEmembership{Member,~IEEE,} Bilal Taha,~\IEEEmembership{Member,~IEEE,} Dimitrios Hatzinakos,~\IEEEmembership{Fellow,~IEEE}
        % <-this % stops a space
\thanks{Ensieh Khazaei, Alireza Esmaeilzehi, Bilal Taha, and Dimitrios Hatzinakos are with the Edward S. Rogers Sr. Department of Electrical and Computer Engineering, University of Toronto, Toronto, Canada (e-mail: ensieh.khazaei@mail.utoronto.ca; alireza.esmaeilzehi@utoronto.ca; bilal.taha@mail.utoronto.ca; dimitris@comm.utoronto.ca).}}% <-this % stops a space
%\thanks{Manuscript received April 19, 2021; revised August 16, 2021.}}

% The paper headers
\markboth{Journal of \LaTeX\ Class Files,~Vol.~14, No.~8, July~2024}%
{Shell \MakeLowercase{\textit{et al.}}: A Sample Article Using IEEEtran.cls for IEEE Journals}

%\IEEEpubid{0000--0000/00\$00.00~\copyright~2021 IEEE}
% Remember, if you use this you must call \IEEEpubidadjcol in the second
% column for its text to clear the IEEEpubid mark.

\maketitle

\begin{abstract}
 In the realm of ubiquitous computing, Human Activity Recognition (HAR) is vital for the automation and intelligent identification of human actions through data from diverse sensors. However, traditional machine learning approaches by aggregating data on a central server and centralized processing are memory-intensive and raise privacy concerns. %Yet, centralizing data for machine learning model training raises concerns like privacy and centralized processing costs. 
 Federated Learning (FL) has emerged as a solution by training a global model collaboratively across multiple devices by exchanging their local model parameters instead of local data. %  without data exchange, instead aggregating model weights. 
 However, in realistic settings, sensor data on devices is non-independently and identically distributed (Non-IID). This means that data activity recorded by most devices is sparse, and sensor data distribution for each client may be inconsistent. As a result, typical FL frameworks in heterogeneous environments suffer from slow convergence and poor performance due to deviation of the global model's objective from the global objective. %traditional FL methods in the heterogeneous environment may incur a drifted global model that causes slow convergence and a heavy communication burden. 
 Most FL methods applied to HAR are either designed for overly ideal scenarios without considering the Non-IID problem or present privacy and scalability concerns. 
 This work addresses these challenges, proposing CDFL, an efficient federated learning framework for image-based HAR. CDFL efficiently selects a representative set of privacy-preserved images using contrastive learning and deep clustering, reduces communication overhead by selecting effective clients for global model updates, and improves global model quality by training on privacy-preserved data. Our comprehensive experiments carried out on three public datasets, namely Stanford40, PPMI, and VOC2012, demonstrate the superiority of CDFL in terms of performance, convergence rate, and bandwidth usage compared to state-of-the-art approaches. %gathered in both controlled and real-life conditions, reveal that our proposed approach outperforms top FL methods, delivering superior results and faster convergence in HAR datasets.

\end{abstract}


\section{Introduction}
In an increasingly connected world, Human Activity Recognition (HAR) has emerged as a pivotal component in the landscape of ubiquitous computing applications~\cite{guo2018multisensor, gaglio2014human, %nikpour2023spatial, 
kamel2018deep}. Of particular interest is the image-based HAR which pertains to the automatic and intelligent identification of actions/behaviors exhibited by humans using image data collected from various sources~\cite{dang2020sensor, esmaeilzehi2024harwe}. These images are often captured by cameras embedded in a wide array of devices, such as smartphones, surveillance systems, and smart environments. Image-based HAR plays a significant role in various domains, including healthcare for patient monitoring, sports analytics, security surveillance systems, and human-computer interaction in smart environments~\cite{zhou2020deep,moniruzzaman2021human, qi2020smartphone}. As these devices continue to proliferate, they generate vast amounts of data with the potential to significantly enhance the accuracy and reliability of HAR algorithms. However, centralizing data from multiple sources on powerful servers or cloud platforms for training machine learning models raise several issues including privacy invasions, potential misuse of sensitive personal data, and the significant costs associated with data transmission and centralized processing.

Federated Learning (FL) has recently emerged as a machine learning paradigm that seeks to address the above-mentioned issues~\cite{McMahan2016}. It allows a global model to be trained across multiple decentralized sources (or clients) holding local data samples, without exchanging them. Instead of transferring raw data to a central server, FL involves training machine learning models locally on each device and then aggregating the model weights (or gradients) to construct an improved global model. This approach ensures that sensitive raw data never leaves the local device, thereby significantly enhancing privacy and security of the users. In the context of HAR, FL is especially promising since it allows for the leveraging of extensive, rich, and diverse images collected from multiple sources employed by different individuals in various environments. Such a decentralized approach does not only preserve privacy but also promises to deliver more personalized and context-aware HAR models by learning from data that is representative of heterogeneous user behaviors. Nonetheless, there are three main challenges in FL that limit its potential: data heterogeneity, communication overhead, and naive aggregation. 

%\begin{itemize}
    %\item 
    %Data heterogeneity: 
    In real-world applications, the data among the clients can be non-IID (independent and identically distributed) which can negatively affect the performance of typical FL system\cite{McMahan2016}. In this scenario, some clients may not contain any sample from some categories of data. %Therefore, the objective of each local model is distant from the global objective \cite{MOON2021}. 
    Recent studies have proposed methods to address this problem. FedProx \cite{FedProx2020} utilizes an $\ell_2$ regularization term to control the distance between the local and global weights. Personalized FL is another approach that addresses the data heterogeneity issue by training personalized models \cite{APFL2020, wang2019, arivazhagan2019, jiang2023, Huang2020, Dinh2020, Fallah2020}. In personalized FL, the focus on local data can prevent the clients from exploiting the information lying in the whole distributed dataset \cite{cFedFCN2023}. Furthermore, some research studies \cite{Sattler2021,Ghosh2022,Chen2020,Briggs2020,Duan2022} employ clustered FL to handle the data heterogeneity. In clustered FL, clients with similar data distributions are grouped into the same cluster, where each cluster collaboratively trains its own global model independent of other clusters.

    %\item 
    %Communication overhead: 
    In every communication round of conventional FL systems, all active clients must send their weights to the server, which poses challenges in situations where there are limitations in bandwidth. Additionally, there are clients with lower convergence rates compared to others, known as stragglers. Therefore, a strategy is required to select clients intelligently not only to reduce communication overhead but also to achieve better performance in a smaller number of communication rounds. Ouyang et al. \cite{Ouyang2021} have proposed cluster-wise straggler dropout and correlation-based client selection techniques to reduce communication overhead for HAR applications. In \cite{Eiffel2022}, an algorithm is proposed to select a subset of devices in resource-constrained environments while achieving the best resource efficiency. The algorithm dynamically selects clients and adaptively adjusts the frequency of model aggregation, considering various factors such as local loss, data size, computational power, resource demands, and the age of updates for each client.

    %\item 
    %Naive aggregation: 
    For updating the server model in a conventional FL framework such as Federated Averaging (FedAvg) \cite{McMahan2016}, a coefficient is assigned to each client according to its number of samples and then clients' model weights are aggregated using weighted averaging. In this method, the global model's objective is far from the global objective for the whole distributed dataset, and it has a negative effect on the convergence rate of the FL system especially in a heterogeneous state. To overcome this issue, the authors in \cite{Chen2023} have proposed a novel aggregation method based on parameter sensitivity. Their method resolves the global model drifting by increasing updates for less sensitive parameters and diminishing updates for parameters with higher levels of sensitivity.

Although existing FL frameworks focus on addressing one or two of these challenges, there is a need to propose a comprehensive FL framework to resolve all of the above challenges. In this work, we propose, CDFL, an efficient federated learning framework for image-based human activity recognition to resolve the above challenges. Our proposed method not only shows an overall improvement in system performance in comparison to state-of-the-art methods but also reduces the communication overhead between the server and various clients.
The contributions of the work of this paper are summarized as follows:

\begin{itemize}
    \item We address the issue of data heterogeneity by introducing a privacy-preserved subset of the dataset to the server. %We address the issue of data heterogeneity by introducing an efficiently selecting subset of the dataset to the server. 
    To choose the best images efficiently, we leverage the ideas of deep clustering and contrastive learning methods. Specifically, before transmitting the data of each client to the server, we manipulate it using image pixelization operation, in order to make it resilient to the leakage of client's sensitive information. %Before transmission to the central server, the selected subset of data is manipulated to be resistant to the leakage of clients' sensitive information. 
    Since the chosen dataset is a good representative of the entire dataset, it effectively overcomes the data heterogeneity challenge.

    \item We reduce the communication overhead by selecting the top-performing clients in each communication round. Notably, in each communication round, the most effective clients share their weights with the server for aggregation. Client selection not only decreases the bandwidth usage in our proposed scheme but also improves the performance because the less effective clients do not contribute to the global model update. 

    \item We improve the quality of the global model by training it on the privacy-preserved subset of data instead of a naive aggregation. In terms of distribution, the selected subset of privacy-preserved images %dataset 
    and the entire dataset are similar. Therefore, training the global model on this subset will approach the global model's objective to the global objective and enhance the system's convergence.
    
    \item We conduct comprehensive experiments on three HAR datasets under different settings and demonstrate that CDFL %our proposed algorithm 
    is able to outperform the state-of-the-art FL schemes existing in the literature in terms of communication efficiency and performance. In particular, we demonstrate that the proposed framework can increase the performance by up to 10\% and speed up the convergence rate by up to 10 times compared to the state-of-the-art approaches. Additionally, it will reduce the bandwidth usage by up to $\sim$64\% in large-scale scenarios.
\end{itemize}
The remainder of this paper is organized as follows. 
In Section \ref{sec:related-work}, we briefly explain previous work related to FL and HAR applications. Section \ref{section:Methodology} explains CDFL in detail. Specifically, we include how we train individual clients locally, choose images using deep clustering, and conduct server training along with client selection. The results of various experimentations on the proposed scheme, as well as, their analysis are described in Section \ref{section:experiments}. %In Section \ref{section:experiments}, we present the results of our experiments and discuss them. 
Finally, the concluding remarks on the work of this paper are made in Section \ref{section:conclusion}.

%\textbf{Explaining the algorithm and summarizing our contribution. In summary, we have three novelties:   }

\section{Related Work}
\label{sec:related-work}

\subsection{Federated Learning } 
FL is a prominent distributed machine learning approach that has attracted attention in recent years. The main goal of FL is to preserve privacy through decentralized learning. The first attempt in this area is FedAvg \cite{McMahan2016} algorithm. Although the results obtained from FedAvg are promising compared to centralized learning, this scheme suffers from several challenges, including data heterogeneity, communication overhead, and naive aggregation. 

To mitigate the data heterogeneity issue, Li \textit{et al.} \cite{MOON2021} presented a model-contrastive federated learning (MOON) to similarize the representation of local models to those of the global model. While MOON performs effectively when the global model is consistent with the global objective, it overlooks the fact that a deviated global model, particularly in strongly heterogeneous conditions, can lead to deviation of all local clients from the global objective. Another category of algorithms focuses on clustered FL to manage heterogeneous states such as cFedFN \cite{cFedFCN2023}, IFCA \cite{Ghosh2022}, FedCluster \cite{FedCluster2020}, FedGroup \cite{FedGroup2021}, FlexCFL \cite{Duan2022}, CFL \cite{Sattler2021}. IFCA \cite{Ghosh2022} is an iterative federated clustering algorithm that alternates between estimating the cluster identities and minimizing the loss function. This method does not need a centralized clustering algorithm, leading to a considerable reduction in server computational cost. %They also mathematically proved
It has been shown that the convergence rate of their proposed algorithm follows the exponential curve. CFL \cite{Sattler2021} is another cluster-based FL framework that does not require any prior knowledge about the number of clusters. In this framework, the geometric properties of the FL loss are leveraged for client clustering. Further, they discovered that by calculating the cosine similarity between client weights, the similarity of their data distribution can be perceived. %Consequently, it leads to a superior clustering outcome, even when there are mild assumptions about the clients and their data distribution.
In cFedFN \cite{cFedFCN2023}, Wei \textit{et al.} have argued that feature norm is a suitable metric to cluster clients in view of its capability to reflect the clients' data distribution. By leveraging feature norms for clustering, a performance improvement is achieved without the need for any pre-clustering steps. The additional computation cost required for feature norm calculation of cFedFN is negligible, which makes it applicable in real-life situations. While clustered FL methods can mitigate data heterogeneity issues by identifying underlying client clusters, clustering becomes challenging as the number of clients increases.%Besides, the additional communication cost required for feature norm calculation in their algorithm is negligible which is important in realistic applications.

A number of existing works~\cite{albaseer2023data} %, zhang2023federated}
have focused on communication overhead reduction to handle bandwidth-limited situations. Chen \textit{et al.} \cite{Chen2021} have designed a probabilistic client selection such that clients that can potentially enhance the convergence rate are selected with higher probability in the following communication round. Besides, in this scheme, a quantization method has been developed to shrink the model parameters before transmission to the server. In \cite{Chen2020ASWT_FedAvg}, an asynchronous model update algorithm is proposed to reduce 
client-server communication. In this algorithm, the various layers of the model are divided into shallow and deep categories. The parameters in the deep category of layers are updated less often compared to the parameters in the shallow layers. With this strategy, the number of parameters to be exchanged between the server and clients is reduced per communication round. Although their proposed strategy is successful in reducing the total communication cost, it needs more communication rounds to achieve a target accuracy compared to FedAvg. The FetchSGD algorithm, presented in \cite{Rothchild2020}, is based on the Count Sketch data structure to overcome the communication bottleneck issue. The FetchSGD sends the sketches of each local client to the server for aggregation. In the server, the momentum and error accumulation are applied to the aggregated sketch. Then, top-k values are selected as sparse updates to broadcast to the connected clients in the following round.  %Furthermore, they provided the mathematical proof for convergence guarantees of their suggested algorithm besides its empirical efficiency on CIFAR10, CIFAR100, and FEMNIST datasets.  

On the server side, naive aggregation of local models suffers from global %client 
drift in heterogeneous situations \cite{Chen2023}, and therefore, deteriorates the FL performance. Chen \textit{et al.} \cite{Chen2020ASWT_FedAvg} have designed a temporally weighted aggregation strategy to resolve this issue. This scheme is based on timestamps, and local clients with recent updates get larger weights compared to other clients. %They observed the positive effect of this method on the accuracy and convergence of the global model. 
FedMA \cite{FedMA2020} has used parameter averaging of the local clients by utilizing a permutation matrix. Specifically, 
%Another algorithm called FedMA \cite{FedMA2020} also deals with the naive aggregation problem. In their work \cite{FedMA2020}, they demonstrated that to achieve meaningful averaging in the parameter space, local weights need to be adjusted using multiplication by a permutation matrix. 
this scheme ensures that similar elements in the model, like channels in convolution layers or neurons in fully connected layers, are matched and averaged appropriately. However, the computational cost of finding the optimal permutation matrix hinders its practical usage. %They used a solver called the Beta-Bernoulli Process-Maximum a Posteriori (BBP-MAP) \cite{Thibaux2007} to calculate this permutation matrix.
EK \textit{et al.} \cite{EK2021} introduced a new aggregation algorithm called FedDist, which follows a similar approach to FedMA but focuses on identifying differences between specific neurons among the clients to modify the global model's architecture. It has been demonstrated that this algorithm considers the clients’ specificity without affecting the overall performance.

%\textbf{I can transfer server aggregation under the category of data heterogeneity.}

%\textbf{The related work for server aggregation should be added}

\subsection{Federated Learning and HAR }
Many existing works in HAR applications still rely on central servers to train their models. This exposes the users' data to substantial privacy threats, especially when dealing with sensitive user information. To maintain user privacy, federated learning has been increasingly adopted in the context of HAR \cite{Zhou2022, EK2021, Concone2022}. %Studies done in the intersection of FL and HAR can be divided into different categories according to their focus. 

Xiao \textit{et al.} \cite{Xiao2021} paid attention to the structure of the model in the FL system. They designed an FL system with an advanced feature extractor for recognizing human activities. This feature extractor consists of two main components. The initial component utilizes convolutional layers to capture local features, while the second component combines LSTM and attention layers to uncover the global relationships within the data.
Another line of studies tries to tackle data heterogeneity in HAR applications. Gad \textit{et al.} \cite{Gad2023} introduced Federated Learning via Augmented Knowledge Distillation (FedAKD) to manage the heterogeneous state. They have demonstrated that FedAKD not only surpasses FedAvg in terms of communication efficiency but also outperforms FedMD \cite{li2019fedmd} due to the incorporation of augmentation techniques. 
FedDL, as presented in Tu et al.'s work \cite{Tu2021}, is another FL framework tailored for activity recognition purposes under non-IID circumstances. FedDL clusters clients based on the similarity of model weights and subsequently integrates models through an iterative layer-wise approach. With this strategy, the system dynamically acquires personalized models for distinct users. Although FedAKD \cite{Gad2023} and FedDL \cite{Tu2021} address the data heterogeneity issue, they do not resolve the communication overhead problem that can be challenging in large scale scenarios. ClusterFL \cite{Ouyang2021} also directed its attention toward client clustering, in which, the cluster indicative matrix that captures the similarity between users, is utilized. %They introduced the new concept of cluster indicator matrix which captures the similarity between users. 
By employing alternating optimization techniques, they iteratively update both local models and the cluster indicator matrix. 

Some research endeavors focus on the FL system in a semi-supervised approach to address the limited availability of labeled samples in the HAR problem. Notably, Bettini \textit{et al.} \cite{Bettini2022} introduced FedHAR which is a semi-supervised FL framework. Specifically, in this scheme, a combination of active learning and label propagation is employed to semi-automatically annotate data for each individual client, effectively addressing the scarcity of labeled samples. %Moreover, transfer learning techniques are exploited to personalize the global model for each client. 
In \cite{zhao2021}, a semi-supervised learning-based HAR scheme is proposed, in which %an activity recognition system is proposed for a semi-supervised state. In this system, 
clients train autoencoders in an unsupervised approach with their unlabeled local data while the server trains an activity classifier through supervised learning. Their scheme %suggested framework 
can achieve competitive performance compared to the supervised approach in the HAR application.

%\textbf{Final conclusion and show the importance of our work. To the best of my knowledge, no work is done on image-based HAR and FL. All the previous works are on signal data.}

\section{Proposed Approuch}
\label{section:Methodology}
In this section, we develop CDFL as an efficient FL framework for HAR. Our proposed scheme addresses the challenges of privacy-preserving and efficient model training in a distributed environment. %CDFL leverages a novel approach to aggregating local model updates with its training while allowing individual clients to adapt their models based on their local data characteristics. 

\subsection{Preliminaries}
In the proposed framework, a given dataset, \(D\), containing \(|C|\) distinct classes, is distributed across \(N\) distinct clients, which \(N\) represents the total number of clients. For any given client, designated as the \(i\)-th party, their local dataset is represented as $D_i=\{ (x_i^j, z_i^j, y_i^j) \}_{j=1}^{n_i}$, where \(x_i^j\) refers to the original image of the \(j\)-th sample in the \(i\)-th client, \(z_i^j\) is the pixelized version of $x^i_j$, where the human identity (face) is covered using image pixelization operation, \(y_i^j\) is the human activity label corresponding to $x^i_j$.
The term \(n_i\) represents the total number of samples that are present within the \(i\)-th client's dataset.% and the total number of samples across all clients is $|D|=\sum\limits_{i=1}^N n_i$.
%\begin{itemize}
%    \item \(x_i^j\) refers to the original image of the \(j\)-th sample in the \(i\)-th client.
%    \item \(z_i^j\) denotes the pixelized variant of the original image.
%    \item \(y_i^j\) is the associated label for that sample.
%\end{itemize}

It's imperative to mention that due to the non-independent and identically distributed (non-iid) nature of our setup, there might be clients, in which certain classes are absent. Therefore, while \(|C|\) represents the total number of classes in dataset \(D\), a client's local dataset might contain fewer classes, represented as \(|C_i|\). This relationship can be mathematically defined as:
\begin{align*}
    |C_i| &\leq |C| \\
    C &= \bigcup_{i=1}^N C_i.
\end{align*}

Furthermore, for each client, two neural networks \(g(.;\theta_i)\) and \({g}(.;\Tilde{\theta}_i)\) are associated with the \textit{personalized local} model and \textit{local} model of the $i$-th client. The functionality of these two networks is explained in the following paragraphs. 

Each of the two neural networks mentioned above consists of two principal segments; 1) a feature extractor that maps the input data (i.e., images in the context of our HAR application) into embedding vectors; 2) a classifier that maps the embedding vector into categorical vectors to make a classification decision.

Let $\theta_i$ and $\Tilde{\theta}_i$ represent, respectively, the weights of \textit{personalized} and \textit{local} models for $i$-th client. %representing the weight vectors of these models by \(\theta\) and \(\Tilde{\theta}\) respectively, 
We denote the penultimate layers of personalized and local models of $i$-th client with \(g^p(.;\theta_i)\) and \({g}^p(.;\Tilde{\theta}_i)\), respectively.
%the outputs from their penultimate layers can be denoted as \(g^p(.;\theta)\) for the personalized model and \({g}^p(.;\Tilde{\theta})\) for the local one.
In fact, $g^p(.;\theta_i)$ and ${g}^p(.;\Tilde{\theta}_i)$ are feature extractors in personalized and local models that provide the representations of the input in the embedding space. For simplicity, in the following sections, the notion ${g}^p(.;\Tilde{\theta_i})$ and ${g}^p(.;{\theta_i})$ is represented as $\Tilde{g}_i^p(.)$ and ${g}_i^p(.)$, respectively.

\subsection{Model Contrastive Learning}
\label{section:contrastive learning}

%As illustrated in Figure \ref{fig:client-training}, 
Each client in our scheme partakes in the training of two distinct neural architectures: the \textit{personalized} model and the \textit{local} model. 

\begin{itemize}
    \item \textbf{Personalized Model:} The objective of this model is to minimize the canonical cross-entropy loss using the original image data. For the $i$-th client's data sample $(x_i^j, z_i^j, y_i^j)$, the personalized model's loss is given as:
    \begin{equation}
        \mathcal{L}_{CE}^{personalized} = CE(g(x_i^j ; \theta_i), y_i^j)
        \label{eq:first}
    \end{equation}
    Utilizing original images poses privacy concerns since they could inadvertently disclose participant identities. To address this, we propose another model for every client, that is trained with the pixelized visual signals. By employing image pixelization, it is assured that the sensitive data in the images, in our case the faces, are masked. Hence, in addition to sending the parameters of each local model to the server, we can transmit their pixelized data (that preserves the human identities) as well. This strategy could have a positive impact on enhancing the FL performance.

    %we introduce a local model for each client that ensures any sensitive data in the images, in our case the faces, are masked. 
    
    To generate the masked images, we employ the RetinaFace model \cite{Retinaface}, an efficient facial detection scheme, to detect faces within these images. Subsequently, the detected faces undergo pixelization through bilinear downsampling and nearest neighbor upsampling operations, both conducted with a factor of $a$, as follows:
    \begin{equation}
        \label{eq:pixelization}
        z_i^j = \text{Nearest Neighbor}_{\uparrow a}(\text{Bilinear}_{\downarrow a}(x_i^j))
    \end{equation}  
    A substantial value for $a$ ensures effective pixelization. Figure \ref{fig:pixelization} demonstrates the face masking process for a sample of Stanford40 dataset \cite{Yao2011}. %\bt{(we need some basic results showing visually, the effect of $a$ on the images.}

    \begin{figure}[t]
    \centering
    \includegraphics[width=0.5
    \textwidth]{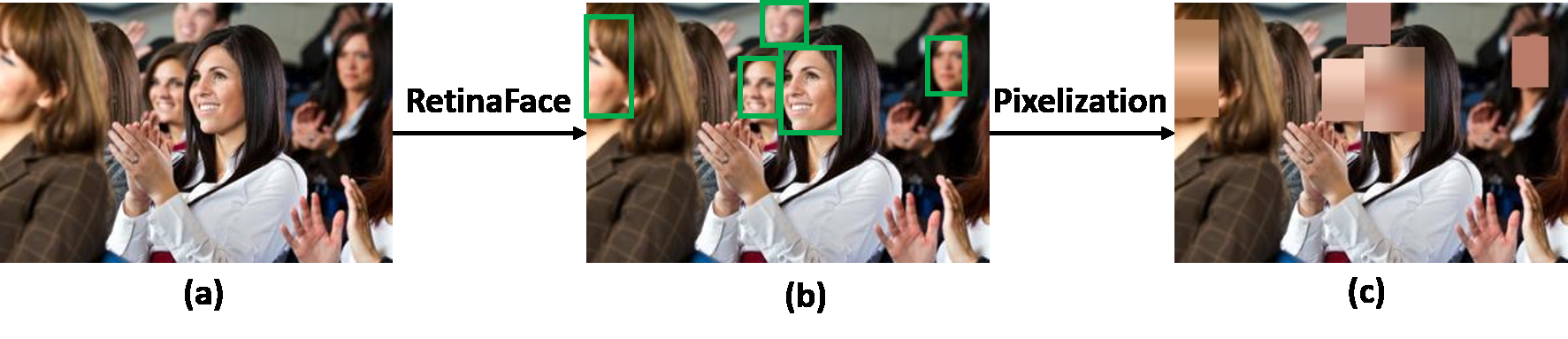}
    \caption{Face masking workflow on a sample of Stanford40 dataset}
    %\caption{Identity protection workflow / Face masking workflow / Face anonymization workflow on a sample of Stanford40 dataset}
    \label{fig:pixelization}
    \end{figure}
   \begin{figure}[tp]
    \centering
    \includegraphics[width=0.48\textwidth]{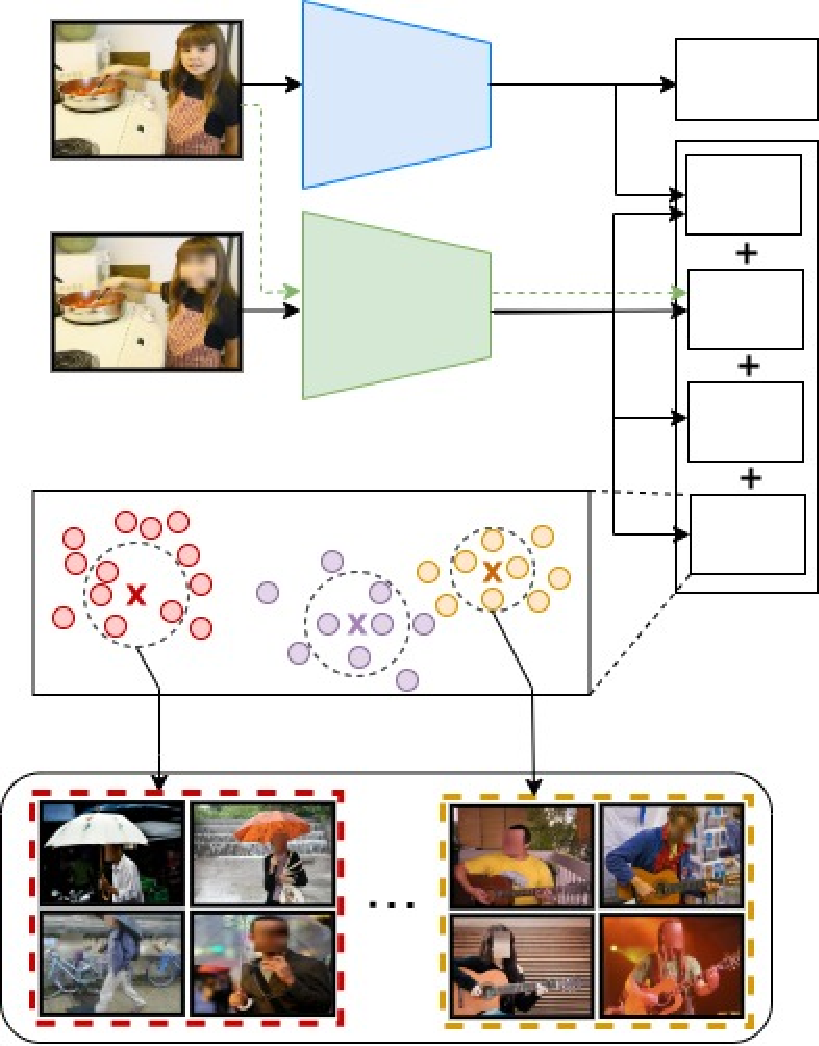}
    \begin{tikzpicture}[overlay, remember picture]
      \node at (-8.6,10.1) {$x_i$}; % Another example formula
      \node at (-8.6,7.9) {$z_i$};

      \node at (-4.7,10.1) {$g_i^p(.;\theta_i)$};
      \node at (-4.7,7.9) {${g}_i^p(.;\Tilde{\theta}_i)$};

      \node at (-2.9,10.45) {$g_i^p(x_i)$};
      \node at (-2.9,7.55) {$\Tilde{g}_i^p(z_i)$};
      \node at (-2.9,8.25) {$\Tilde{g}_i^p(x_i)$};

      \node at (-0.9,10.3) {$\mathcal{L}_{CE}^{personal}$};

      \node at (-0.9,9.0) {$\mathcal{L}_{KD}$};
      \node at (-0.9,7.8) {$\mathcal{L}_{con}$};
      \node at (-0.9,6.6) {$\mathcal{L}_{CE}^{local}$};
      \node at (-0.9,5.4) {$\mathcal{L}_{cl}$};

      \node at (-7.4,6.1) {Image Selection};

      \node at (-8.0,3.1) {$\mathcal{S}_i$};
      
    \end{tikzpicture}
    \caption{Overview of local training and image selection within the $i$-th client}
    \label{fig:client-training}
\end{figure}
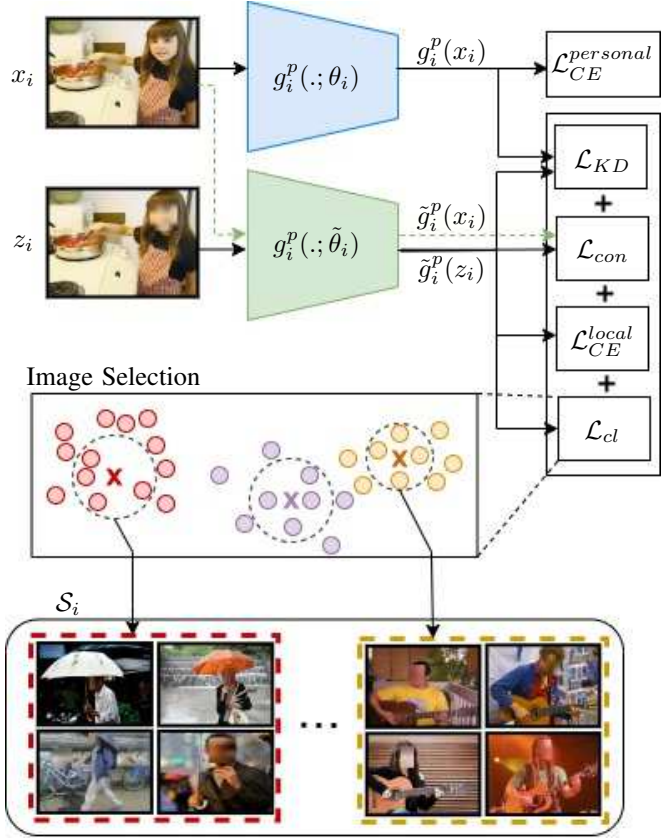
    \item \textbf{Local Model:} The local model is trained on pixelized images using a similar architecture as the personalized model. Its design addresses the aforementioned privacy concerns, aiming to achieve reasonable performance on pixelized images. The loss for this model is the standard cross-entropy based on pixelized images, which is represented as
    \begin{equation}
        \mathcal{L}_{CE}^{local} = CE({g}(z_i^j ; \Tilde{\theta}_i ), y_i^j)
        \label{eq:second}
    \end{equation}
    
    The proposed local model incorporates both knowledge distillation and contrastive learning to improve the model's capability to generate efficient representations for pixelized images, as shown in Figure \ref{fig:client-training}. 
    With the addition of contrastive learning, the local model aims for congruent representations between original and pixelized images. Similar to \cite{Hook2016}, we define the contrastive loss as:
    \begin{equation}
        \mathcal{L}_{con} = - \log\frac{e^{\frac{cos(\Tilde{g}_i^p(z_i^j), \Tilde{g}_i^p(x_i^j))}{\tau}}}{e^{\frac{cos(\Tilde{g}_i^p(z_i^j), \Tilde{g}_i^p(x_i^j))}{\tau}} + \sum_{k \neq j} e^{\frac{cos(\Tilde{g}_i^p(z_i^j), \Tilde{g}_i^p(x_k^j))}{\tau}}}
        \label{eq:third}
    \end{equation}
    %For simplicity, the notion ${g}^p(.;\Tilde{\theta_i})$ is represented as $\Tilde{g}_i^p(.)$. 
    Here, $\Tilde{g}_i^p(z_i^j)$ and $\Tilde{g}_i^p(x_i^j)$ represent the local model's representations over pixelized and original images respectively. The term $\tau$ denotes the temperature parameter that controls the strength of penalties on hard negative samples, while $cos(., .)$ is the cosine similarity metric. %It is defined between vectors $\textbf{a}$ and $\textbf{b}$ as $cos(\textbf{a}, \textbf{b}) = \frac{\textbf{a} \cdot \textbf{b}}{\| \textbf{a} \| \| \textbf{b} \|}$.
    
    To further refine the local model's performance on original images, knowledge distillation is employed, leveraging insights from the personalized model which enhances the similarity between the local model representation on pixelized images and the representations of the personalized model on the original counterparts:
    \begin{equation}
        \mathcal{L}_{KD} = - \log\frac{e^{\frac{cos(\Tilde{g}_i^p(z_i^j), {g}_i^p(x_i^j))}{\tau}}}{e^{\frac{cos(\Tilde{g}_i^p(z_i^j), {g}_i^p(x_i^j))}{\tau}} + \sum_{k \neq j} e^{\frac{cos(\Tilde{g}_i^p(z_i^j), {g}_i^p(x_k^j))}{\tau}}}
        \label{eq:forth}
    \end{equation}
    where ${g}_i^p(x_i^j)$ %stands for ${g}^p(x_i^j ; {\theta_i})$, 
denotes the personalized model's representations over the original images.
    
    Finally, the total loss used for training the local model is as:
    \begin{equation}
        \mathcal{L}^{local} = \mathcal{L}_{CE}^{local} + \lambda_1 \mathcal{L}_{con} + \lambda_2 \mathcal{L}_{KD}
        \label{eq:fifth}
    \end{equation}
    here, $\lambda_1$ and $\lambda_2$ work as hyperparameters to calibrate the contributions from contrastive learning and knowledge distillation components, respectively.
\end{itemize}

\subsection{Deep Clustering for Image Selection} \label{section:deep clustering}
Clustered Federated Learning (FL) \cite{Xu2023, Shlezinger2020,Chengxi2022, Kim2021} categorizes clients with similar data distributions into groups, and consequently, trains a distinct global model for every cluster. %The effectiveness of clustered FL hinges on a pertinent clustering criterion. 
Previous investigations have adopted diverse criteria for client clustering, which range from using contextual information \cite{FedCluster2020}, model parameters \cite{Sattler2021,Xueyang2021, FedGroup2021,Duan2022}, loss functions \cite{Ghosh2022}, and embedding vectors \cite{cFedFCN2023, Jamali2022}.
While earlier efforts largely focused on client clustering to mediate data heterogeneity, our research harnesses image clustering to transmit a minimal number of representative pixelized images to the server in order to enhance the FL performance. It should be noted that local data remains unshared with servers or other clients in conventional FL systems. %This restriction leads to slowed convergence and subpar performance. To effectively handle both privacy concerns and data scarcity, our proposed novel solution entails transmitting pixelized rather than original images to the server. 
As delineated in Section \ref{section:contrastive learning}, pixelized images effectively obscure individual sensitive data and contain the necessary information for performing the task of HAR. Nonetheless, the indiscriminate transmission of all pixelized images is memory-consuming and inefficient, emphasizing the need for efficient image selection methodologies.

%To reflect client-specific characteristics, pixelized images must be practically chosen. We deploy deep clustering, operating within the embedding space, for optimal image selection.
In order to perform image selection, we utilize the idea of deep clustering in the feature space. Mathematically, deep clustering for the $i$-th client is expressed as: 

\begin{align}
    \sum\limits_{k=1}^{|C_i|} \mathbbm{1}_{jk} \| \Tilde{g}_i^p(z_i^j) - \mu_{ik}\| 
    \label{eq:sixth}
\end{align}
where $\mu_{ik}$ is the centroids corresponding to the $k$-th cluster in the $i$-th client. The centroids will be learned during training of the local model. $\mathbbm{1}_{jk}$ is an indicator that becomes one if $\Tilde{g}_i^p(z_i^j)$ belongs to the $k$-th cluster, and zero otherwise. 

The training parameters in Equation (\ref{eq:sixth}) scale with the embedding space's size. For obtaining high-quality centroids for each client, their sample count should be considerably large relative to this space. Given that FL systems rarely assign individual clients substantial samples, the deep clustering cannot be carried out in an efficient manner \cite{Dang2021, Dong2022, Barlow2021, Lv2021}. In order to address this, we employ the pseudo-labeling \cite{Niu2022, Zhang2021, Wang2021} technique for carrying out the deep clustering operation. %This prevalent deep clustering strategy enables convergence to salient cluster configurations, selecting model weights as the training parameters, rather than cluster centroids. It involves generating pseudo-labels for data samples using the predictions of a model trained on the existing labeled data. These pseudo-labels are then considered as the actual ground truth labels, and the model is re-trained on the data samples. 
In this prevalent deep clustering strategy, first, cluster centroids are derived through methods like k-means clustering. These centroids serve as the reference points for assigning data samples to their corresponding clusters. These assigned cluster labels are then considered as the samples' pseudo-labels. Subsequently, the model is re-trained with the objective of aligning the representations of the samples with their corresponding cluster centroids.

Our algorithm employs the global model in conjunction with K-means for pseudo-label prediction, leveraging the global model's encompassing ability to derive representations from all clients. In the first stage, the global model %processes 
is applied to pixelized images within each client, yielding individual representations in the embedding or penultimate layer for each client:
\begin{align}
    g(\{z_i^j\}_{j=1}^{n_i};\Tilde{\theta}) = \{\Tilde{g}^p(z_i^j)\}_{j=1}^{n_i}
    %\{z_i^j\}_{j=1}^{n_i} \xrightarrow{g(.;\Tilde{\theta})} \{\Tilde{g}^p(z_i^j)\}_{j=1}^{n_i}
    \label{eq:seventh}
\end{align}

Subsequently, K-means extracts cluster centroids and pseudo-labels, given the representations learned earlier as follows: 
\begin{align}
    \{ \Tilde{\mu}_{ik} \}_{k=1}^{|C_i|} =\arg\min\limits_{\{ \Tilde{\mu}_{im} \}_{m=1}^{|C_i|}} \sum\limits_{j=1}^{n_i} a_{jm}^i \| \Tilde{g}^p(z_i^j) - \Tilde{\mu}_{im} \| \
    %\{\Tilde{g}^p(z_i^j)\}_{j=1}^{n_i} \xrightarrow{\text{K-means}} \{ a_{jk}^i \}_{j=1}^{n_i},~ \{ \Tilde{\mu}_{ik} \}_{k=1}^{|C_i|} 
    \label{eq:eighth}
\end{align}
Then the pseudo-labels $a_{jk}^i$ is defined as
\begin{align}
    a_{jk}^i= \begin{cases}
        1 & \text{if }k=\min\limits_{m}  \| \Tilde{g}^p(z_i^j) - \Tilde{\mu}_{im} \| \\
        0 & \text{otherwise}
    \end{cases}
    \label{eq:ninth}
\end{align}
Therefore, Equation (\ref{eq:sixth}) can be reformulated as
\begin{align}
    \mathcal{L}_{cl}=\sum\limits_{k=1}^{|C_i|} a_{jk}^i \| \Tilde{g}_i^p(z_i^j) - \Tilde{\mu}_{ik}\| 
    \label{eq:tenth}
\end{align}
By introducing the deep clustering loss into Equation (\ref{eq:fifth}), the local model's loss can be computed as:
\begin{align}
    \mathcal{L}^{local}= \mathcal{L}_{CE}^{local} + \lambda_1  \mathcal{L}_{con} + \lambda_2 \mathcal{L}_{KD} + \lambda_3 \mathcal{L}_{cl}
    \label{eq:eleventh} 
\end{align}
where $\lambda_3$ adjusts the significance of the clustering term.

After the local model's training using Equation (\ref{eq:eleventh}), K-means determine updated centroids based on the representations generated by the local model: 
\begin{align}
    \{ \bar{\mu}_{ik} \}_{k=1}^{|C_i|} =\arg\min\limits_{\{ \bar{\mu}_{im} \}_{m=1}^{|C_i|}} \sum\limits_{j=1}^{n_i} a_{jm}^i \| \Tilde{g}_i^p(z_i^j) - \bar{\mu}_{im} \| \
    %\{\tilde{g}_i^p(z_i^j)\}_{j=1}^{n_i} \xrightarrow{\text{K-means}} \{ \bar{\mu}_{ik} \}_{k=1}^{|C_i|} 
    \label{eq:twelfth}
\end{align}

Subsequently, for each centroid $\bar{\mu}_{ik}$, the Euclidean distance to all local samples is computed, and the $m$ nearest samples to each centroid are then selected as the optimal samples for transmission to the server. The collection of chosen images for the $k$-th cluster is denoted as:
\begin{align}
    \mathcal{S}_i^k= \{ (z_i^j,y_i^j) \}_{j=1}^m    
    \label{eq:thirthinth}
\end{align}
By performing the selection process on all centroids for each client, a set containing the best images for transmission to the server is obtained as:
%Furthermore, the selection process is performed across all centroids for each client, forming a set:
\begin{align}
    \mathcal{S}_i=\bigcup_{k=1}^{|C_i|}\{ \mathcal{S}_i^k \}%_{k=1}^{|C_i|}    
    \label{eq:forthinth}
\end{align}
where $\mathcal{S}_i$ will be shared with the server for further processing. %It is worth mentioning that $\mathcal{S}_i$ will be stored on the server until the next communication round when client $i$ is selected to contribute to the server update process and update its $\mathcal{S}_i$. Therefore, in each communication round, privacy-preserved data samples from all clients exist on the server independent of clients' contribution.
It is worth mentioning that $\mathcal{S}_i$ remains stored on the server until the next communication round when the respective client is selected to contribute to the server update process and update its corresponding set, $\mathcal{S}_i$. Consequently, in each communication round, privacy-preserved data samples from all clients are present on the server, regardless of their contributions.

The advantages of the proposed image selection method are two-fold. Firstly, it employs global model representations for obtaining the pseudo-labels, and hence, provides centroids with a global view from all clients. This prevents the deviation of the local model from the global model during local model training. Secondly, the global and local models are updated in each communication round. As a result, the selected images are dynamically refined to improve performance over time.

\subsection{Server Training and Client Selection in Federated Learning} 
In traditional FL systems, during each communication round, the central server aggregates model weights from all connected clients to update the global model. There exist three issues with using this approach. First, due to the data heterogeneity among clients, the aggregated global model is far from the global objective. Second, it necessitates significant memory and bandwidth resources, which might be unattainable in many real-world scenarios. Third, some clients may demonstrate slower convergence compared to others. Including such clients during the global model update can detrimentally impact the convergence rate of the entire FL framework. 
\begin{figure}[htp]
  \begin{minipage}[b]{0.23\textwidth}
    %\captionsetup{labelformat=empty}
    \centering
    \includegraphics[scale=0.43]{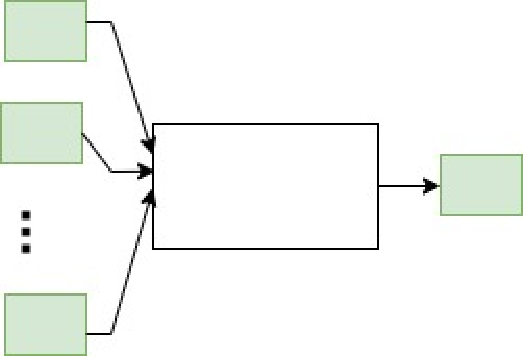}
    \caption*{(a) Server Initialization}
    \label{fig:algorithm-client}
    % Add TikZ overlays for this image here
    \begin{tikzpicture}[overlay, remember picture]
      \node at (-1.6,3.3) {$\Tilde{\theta}_1$};
      \node at (-1.6,2.55) {$\Tilde{\theta}_2$};
      \node at (-1.55,1.15) {$\Tilde{\theta}_N$};

      \node at (0.0,2.2) [font=\fontsize{9.0}{15}\selectfont] {Aggregation};

      \node at (1.6,2.15) {$\Tilde{\theta}$};
    \end{tikzpicture}
  \end{minipage}
  \begin{minipage}[b]{0.23\textwidth}
    %\captionsetup{labelformat=empty}
    \centering
    \includegraphics[scale=0.45]{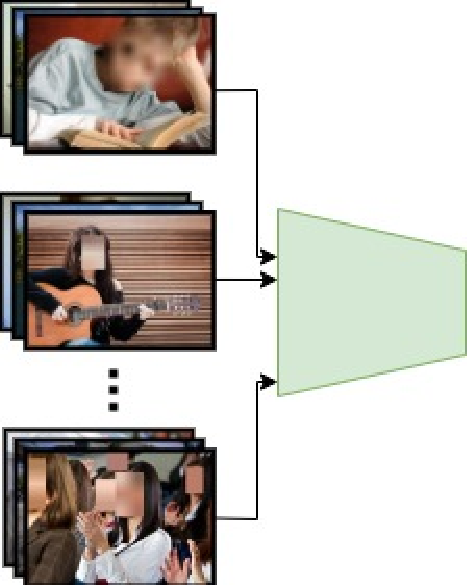}
    \caption*{(b) Server Training}
    \label{fig:algorithm-server}
    % Add TikZ overlays for this image here
    \begin{tikzpicture}[overlay, remember picture]

      \node at (1.0,3.1) {${g}(.;\Tilde{\theta})$};

      \node at (-2.0,4.5) {$\mathcal{S}_1^{tr}$};
      \node at (-2.0,3.05) {$\mathcal{S}_2^{tr}$};
      \node at (-2.0,1.2) {$\mathcal{S}_N^{tr}$};
    \end{tikzpicture}
  \end{minipage}
  \caption{Server update process at each communication round of CDFL}
  \label{fig:server-update}
\end{figure}
In order to address these challenges, we consolidate the global model after naive aggregation using the privacy-preserved selected images. Additionally, we propose a client selection algorithm that identifies and integrates contributions from the most useful clients in each round. This algorithm leverages the dataset of selected images from Section \ref{section:deep clustering} to determine the optimal clients.

For each client, the received dataset, denoted as $\mathcal{S}_i$, is divided into two sets of training ($\mathcal{S}_i^{tr}$) and validation ($\mathcal{S}_i^{val}$). The aggregation of local models follows the fundamental aggregation:
\begin{align} \Tilde{\theta} = \sum\limits_{i \in \mathcal{C}} \frac{n_i}{\sum\limits_{i \in \mathcal{C}} n_i} \Tilde{\theta_i} \label{eq:fifteenth} \end{align}
Here, $\mathcal{C}$ represents the set of selected clients forwarding their model weights to the central server. The process of selecting clients will be explained later in this section.

%The global model achieved through basic aggregation does not reflect the overall objective, mainly because it does not take the data samples into account.
A simple parameter aggregation in the server is not sufficient to approach the global objective, since the global model requires an efficient fine-tuning using the selected privacy-preserved data of all clients in order to align with the global objective and provide a superior FL performance. To mitigate this, we utilize images chosen through the methodology detailed in Section \ref{section:deep clustering} to adjust the global weights, aligning them with the global objective. 

After the naive aggregation of the global model, it undergoes training on the combined training set, $\mathcal{S}^{tr}=\bigcup\limits_{i=1}^N \mathcal{S}_i^{tr}$, and is fine-tuned using the privacy-preserved data samples stored on the server from all clients. %received from all the clients. 
The loss for the global model, for any data instance $(z_i^j, y_i^j) \in \mathcal{S}^{tr}$, is:

\begin{align}
    \mathcal{L}_g = CE({g}(z_i^j ; \Tilde{\theta}),y_i^j)
    \label{eq:sixteenth}
\end{align}
This loss undergoes updates via the Stochastic Gradient Descent (SGD) optimizer. Finally, the updated global model is broadcast to the clients in order to initialize the \textit{personalized} and \textit{local} models for the next communication round. %:
%\begin{align}
%    \Tilde{\theta} \longleftarrow \Tilde{\theta} - \eta_s \nabla_{\Tilde{\theta}} \mathcal{L}_g
%    \label{eq:seventeenth}
%\end{align}
%where $\eta_s$ is the learning rate on the server side. 
This strategy will be helpful to overcome the slow convergence in the typical FL algorithms. The overall architecture of the global model update is shown in Figure \ref{fig:server-update}. 

To optimize the bandwidth, the trained global model is evaluated on the validation subsets and then clients are ranked based on their validation performance in a descending order. A certain percentage of clients %\bt{How is this percentage selected?} 
with the highest validation accuracy are selected to contribute their weights in the next communication round. These selected clients are stored in $\mathcal{C}$ set.
It should be noted that in the next round, the updated global weights are sent to all connected clients. However, after local training is done, only the chosen clients share their weights with the server and update their pixelized images on the central server.

%Moreover, to reduce bandwidth usage, clients are sorted according to the performance of the global model after training on their validation subset. $p$ percent of the whole clients with the best validation accuracy are chosen to share their weights in the following round.

\section{Experiments}
\label{section:experiments}
We evaluate the performance of CDFL on several image-based HAR datasets and compare it with the state-of-the-art frameworks FedAvg~\cite{McMahan2016}, SCAFFOLD~\cite{Karimireddy2020}, FedProx~\cite{FedProx2020}, MOON~\cite{MOON2021}, FedDyn~\cite{Durmus2021}, and FedHKD~\cite{chen2023best}. %We analyze our proposed algorithm experimentally from four aspects: (i) performance (accuracy) comparison of our algorithm versus the SOTA baselines, (ii) the scalability of our proposed algorithm for a large number of clients, (iii) the effect of data heterogeneity, and (iv) cost and convergence rate.

Our experimental analysis of CDFL focuses on five key aspects:
(i) comparing the accuracy of CDFL to state-of-the-art baselines, (ii) evaluating the CDFL's ability to handle a large number of clients, (iii) investigating how CDFL performs in different levels of heterogeneity, (iv) assessing the CDFL's efficiency and convergence speed, and (v) analyzing the impact of different of important parameters on the performance of CDFL.

These evaluations will provide insights into the effectiveness and suitability of CDFL for image-based HAR tasks.

\subsection{Experimental Settings}
\subsubsection{Datasets} Three image-based HAR datasets are used in our experiments, namely Stanford40 \cite{Yao2011}, PPMI \cite{Yao2010}, and VOC2012 \cite{VOC2012}. Stanford40 contains 40 categories of human actions and a total of 9,532 images that are split with ratios of 0.9 and 0.1 as the training and testing sets, respectively. The PPMI includes 4,800 images of people interacting with 12 different musical instruments, and VOC2012 contains 10 classes of actions and 3,744 images. It should be noted that images with more than one action are omitted in VOC2012 to avoid model confusion. PPMI and VOC2012 are split to train and test sets by the original publisher and we follow their partitioning. We use Dirichlet distribution to generate the non-IID data partitions among clients \cite{MOON2021, chen2023best, Durmus2021}. Specifically, we distribute samples among clients with $Dir_N(\alpha)$, where $\alpha$ is the concentration parameter of Dirichlet distribution. Smaller values of $\alpha$ result in a more unbalanced data distribution among clients. We set $\alpha$ to 1.0 in all the experiments unless explicitly specified. 

\subsubsection{Baseline Methods}
To assess the effectiveness of CDFL, we conduct a comparative analysis with six state-of-the-art baselines, including FedAvg \cite{McMahan2016}, SCAFFOLD \cite{Karimireddy2020}, FedProx \cite{FedProx2020}, MOON \cite{MOON2021}, FedDyn \cite{Durmus2021}, and FedHKD \cite{chen2023best}. We also compare with the SOLO approach, where each client trains a model with its local data independent of other clients. In the SOLO approach, no server model is utilized.
%\begin{itemize}
%    \item \textbf{FedAvg} trains local models for each client and then aggregates the clients' weights in the server. The updated global weights are broadcast to the clients.
%    \item \textbf{SCAFFOLD} defines the control variates concept (variance reduction) to correct the client-drift in local updates. In this algorithm, both weights and control variates are communicated between clients and the server. 
%    \item \textbf{FedProx} incorporates $\ell_2$ regularization to minimize the distance between the client and global weights.
%    \item \textbf{MOON} uses contrastive learning techniques to maximize the similarity between local and global model representations in each round while minimizing the similarity between representations of a local model in consecutive rounds. 
%    \item \textbf{FedDyn} employs dynamic regularization term instead of statistic regularizer for each client at each round in order to improve the performance.
%    \item \textbf{FedHKD} leverages clients' hyper-knowledge (the means of clients' representations and the corresponding soft predictions) information to tackle data heterogeneity issue.

%\end{itemize}
%Our implementation is done in PyTorch and our code is available at \textbf{the GitHub address should be provided}.\\
%\textbf{We have the opportunity to remove the explanation of each baseline if we want}
\subsubsection{Hyperparameters} We use the Stochastic Gradient Descent (SGD) optimizer with a constant learning rate of 0.01 across all settings. For the SGD optimizer, we set the weight decay to 0.00001 and the momentum to 0.9. The batch size is uniformly set to 32 for all experiments. In the case of SOLO, the number of local epochs is fixed to 100. For all federated learning approaches, the number of local epochs is set to 5.
The performance evaluation is conducted after 20 communication rounds for the Stanford40 dataset and 25 communication rounds for both the PPMI and VOC2012 datasets. We chose these specific numbers of communication rounds because we observed that considerable improvement is not seen beyond these points. We employ the ResNet-50 as the deep network architecture for all experiments, and in each communication round, 80\% of clients are randomly involved.

In the FedProx algorithm, we choose the best value for $\mu$ from the set of \{0.001, 0.01, 0.1, 1.0\}. For MOON, we tune $\mu$ over the range of \{0.01, 0.1, 1, 5\} and report the best result. The $\alpha$ parameter in FedDyn is adjusted with values from \{0.001, 0.01, 0.1, 1.0\}. In FedHKD, we tune the parameters $(\lambda, \gamma)$ using \{(0.001, 0.001), (0.01, 0.01), (0.05, 0.05), (0.1, 0.1)\}, and report the best result.

In CDFL, 50\% of clients ($|\mathcal{C}|=0.5N$) are selected to share their weights with the server and update their pixelized images on the server. $\lambda_1$, $\lambda_2$, and $\lambda_3$ are set to 0.1, 0.1, and 0.05, respectively. The parameter $m$, the number of nearest selected samples to each centroid, is set to 3 for experiments involving 30 clients and 5 for experiments with 20 clients. When the number of clients is 10, the value of $m$ is 9 for the Stanford40 dataset and 7 for the PPMI and VOC2012 datasets. %Further, in all experiments, the number of selected clients ($|\mathcal{C}|$) is set to $0.5N$, where $N$ is the total number of clients.

\subsection{Performance Comparison}
At the end of each communication round, the local models are applied to the
test samples to evaluate their individual performance. Then, the performance of all connected clients is averaged to obtain the performance of the scheme. The mean accuracy of all schemes, tested with varying numbers of clients (10, 20, and 30), is summarized in Table \ref{tab:performance comparison}. First, we have to say CDFL is the best. The SOLO approach consistently exhibits the poorest performance across all settings, representing the beneficial impact of the FL framework in decentralized learning. As the number of clients increases, there is a drop in performance across all schemes.
In the case of FedAvg~\cite{McMahan2016}, this drop is particularly significant, ranging from 7\% to 16\% among different datasets. %Other baselines add regularization terms to their local loss function to mitigate data heterogeneity and this performance gap. 
The relatively lower accuracy of FedAvg can be attributed to its limited ability to handle heterogeneous conditions. The performance improvement of CDFL in comparison to MOON \cite{MOON2021} is up to 4.2\%, 7.1\%, and 9.8\% for Stanford40, PPMI, and VOC2012, respectively. Among the baselines, the performance of FedDyn~\cite{Durmus2021} is closer to CDFL. Compared to FedDyn~\cite{Durmus2021}, CDFL improves the performance up to 1.3\%, 5.4\%, and 5.7\% in Stanford40, PPMI, and VOC2012, respectively. 
%FedHKD~\cite{Chen2023}, which leverages hyper-knowledge for addressing data heterogeneity issues, does not perform as well as our expectation in the context of HAR datasets, while it shares more knowledge among clients. %This might be due to the greater complexity of the images compared to those found in the original paper \cite{Chen2023}, including SVHN and CIFAR10/100 datasets or the presence of outliers in the embedding space, causing deviations in hyper-knowledge vectors.

As shown in Table \ref{tab:performance comparison}, CDFL consistently outperforms all other baselines across various settings, demonstrating the strength of CDFL in comparison to previous works. In general, CDFL has a smaller standard deviation compared to other baselines. This implies that it not only works well on the whole set of clients according to its higher mean accuracy but also performs well on all individual clients due to its low standard deviation. In contrast, the higher standard deviation in other baselines represents difficulties in achieving good performance on some clients due to non-IID data distribution. Furthermore, CDFL is still able to provide higher performance, even when the number of clients increases, which is crucial in realistic applications where the number of clients is often large.

%The average top-1 accuracy of all schemes with three different numbers of clients 10, 20, and 30 is shown in Table \ref{tab:performance comparison}. The SOLO approach has the worst performance in all settings, representing the positive impact of the FL framework. As the number of clients increases the schemes' performance drop slightly. In FedAvg, this drop is significant, between 7\% to 16\% among different datasets, while other schemes that has regularization try to diminish this gap. Among the FL baselines, FedAvg has low accuracy because it cannot handle heterogeneous conditions. Among the baselines, the performance of FedDyn is closer to the approach. Although FedHKD employed hyper-knowledge to tackle the data heterogeneity issue, its performance on HAR datasets is not satisfactory. The main reason may be the more complex images compared to SVHN and CIFAR10/100 datasets, being used in the original paper or the existence of outliers in the embedding space which results in deviation in hyper-knowledge vectors.
%As shown in Table \ref{tab:performance comparison}, our proposed model consistently outperforms the other baselines in all settings. Further, it tries to keep the performance as high as possible when the number of clients increases, which would be very beneficial in realistic applicaitons.
%\textbf{Report my results for 10 clients of all methods under noniid condition with alpha=1.0}

\begin{table*}[ht!]
  \centering
  \caption{Mean accuracies (\%± standard deviation) for all datasets}
  \label{tab:performance comparison}
  \resizebox{\textwidth}{!}{\begin{tabular}{cccccccccccc}  
    \toprule
    {} & \multicolumn{3}{c}{Stanford40} & {} & \multicolumn{3}{c}{PPMI} & {} & \multicolumn{3}{c}{VOC2012} \\
    \cmidrule{2-4} 
    \cmidrule{6-8}
    \cmidrule{10-12}
        & $N=10$ & $N=20$ & $N=30$ & & $N=10$ & $N=20$ & $N=30$ & & $N=10$ & $N=20$ & $N=30$  \\
    \midrule
    SOLO & $78.42 \pm 8.81$ & $58.32 \pm 3.28$ & $52.46 \pm 4.56$ &   & $67.46 \pm 13.96$ & $54.55 \pm 13.27$ & $47.42 \pm 13.90$ &   & $59.82 \pm 4.14$ & $52.58 \pm 8.20$ & $44.40 \pm 8.05$ \\
    \midrule
    FedAvg \cite{McMahan2016} &  $86.18 \pm 5.07$ & $81.22 \pm 3.50$ & $79.74 \pm 5.34$ &   & $88.22 \pm 3.35$ & $77.78 \pm 14.93$ & $75.60 \pm 11.88$ &   & $74.03 \pm 4.82$ & $67.20 \pm 12.38$ & $65.91 \pm 10.40$ \\
    \midrule
    SCAFFOLD \cite{Karimireddy2020} &  $83.81 \pm 2.80$ & $82.06 \pm 2.00$ & $82.01 \pm 4.79$ &   & $88.31 \pm 2.67$ & $81.74 \pm 14.71$ & $82.26 \pm 12.26$ &   & $70.07 \pm 4.34$ & $71.06 \pm 6.23$ & $73.48 \pm 6.51$ \\
    \midrule
    FedProx \cite{FedProx2020} &  $86.44 \pm 4.05$ & $81.41 \pm 3.09$ & $79.92 \pm 7.51$ &   & $88.13 \pm 4.63$ & $80.21 \pm 9.68$ & $72.94 \pm 15.28$ &   & $75.62 \pm 6.64$ & $72.5 \pm 13.59$ & $70.32 \pm 9.37$ \\
    \midrule
    MOON \cite{MOON2021} &  $87.00 \pm 3.35$ & $81.73 \pm 3.86$ & $80.13 \pm 5.74$ &   & $86.97 \pm 7.39$ & $81.24 \pm 10.78$ & $77.91 \pm 13.66$ &   & $72.39 \pm 9.27$ & $66.26 \pm 16.19$ & $67.02 \pm 9.44$ \\
    \midrule
    FedDyn \cite{Durmus2021} &  $87.02 \pm 3.48$ & $83.79 \pm 2.76$ & $83.04 \pm 5.12$ &   & $88.75 \pm 2.65$ & $82.58 \pm 13.06$ & $81.19 \pm 13.15$ &   & $74.24 \pm 4.07$ & $73.05 \pm 7.88$ & $71.13 \pm 10.88$ \\
    \midrule
    FedHKD \cite{chen2023best} &  $86.59 \pm 5.39$ & $82.31 \pm 4.66$ & $80.04 \pm 10.23$ &   & $89.69 \pm 6.06$ & $83.11 \pm 6.97$ & $73.86 \pm 18.11$ &   & $69.70 \pm 10.32$ & $69.28 \pm 11.78$ & $67.86 \pm 11.06$ \\
    \midrule
    CDFL &  $\mathbf{88.19 \pm 3.65}$ & $\mathbf{84.48 \pm 3.12}$ & $\mathbf{84.37 \pm 2.41}$ &   & $\mathbf{90.74 \pm 2.98}$ & $\mathbf{88.00 \pm 4.66}$ & $\mathbf{85.14 \pm 8.39}$ &   & $\mathbf{77.82 \pm 2.49}$ & $\mathbf{75.19 \pm 8.82}$ & $\mathbf{76.89 \pm 5.71}$ \\
    \bottomrule
  \end{tabular}}
\end{table*}

\subsection{Scalability}
%\begin{table}[b]
% \centering
%  \caption{MEAN ACCURACIES (\%± STANDARD DEVIATION) FOR LARGE NUMBER OF CLIENTS ON STANFORD40}
%  \label{tab:scalability}
%  \begin{tabular}{ccc}  
%    \toprule
%    {} & \multicolumn{2}{c}{Stanford40}  \\
%    \cmidrule{2-3} 
    
%        & $N=50$ &  $N=100$   \\
%    \midrule
%    SOLO &  $45.04 \pm 8.99$ &  $42.24 \pm 6.88$ \\
%    \midrule
%    FedAvg \cite{McMahan2016} &  $75.81 \pm 8.35$ &  $71.82 \pm 6.87$ \\
%    \midrule
%    SCAFFOLD \cite{Karimireddy2020} &  $79.52 \pm 7.93$ &  $80.97 \pm 8.29$  \\
%    \midrule
%    FedProx \cite{FedProx2020} &  $76.14 \pm 7.98$ &  $72.38 \pm 7.21$ \\
%    \midrule
%    MOON \cite{MOON2021} &  $76.93 \pm 7.34$ &  $72.89 \pm 6.73$ \\
%    \midrule
%    FedDyn \cite{Durmus2021} &  $81.89 \pm 7.78$ & $81.77 \pm 10.67$ \\
%    \midrule
%    FedHKD \cite{Chen2023} &  $78.18 \pm 6.07$ &  $73.96 \pm 5.59$ \\
%    \midrule
%    Ours &  $\mathbf{82.55 \pm 5.64}$ &  $\mathbf{82.67 \pm 3.54}$ \\
%    \bottomrule
%  \end{tabular}
%\end{table}

To demonstrate the scalability of CDFL to a large number of clients, we evaluate the Stanford40 dataset in two scenarios with 50 and 100 clients. The data samples were distributed among clients using Dirichlet distribution with $\alpha=1.0$. It is worth noting that Stanford40 is selected for these large-scale scenarios due to the small size of other datasets, whose number of samples is approximately 50\% of the samples available in Stanford40. %Consequently, they are not suitable for the scalability evaluation. 
Since the number of clients increases, the number of communication rounds is set to 30 to make sure that all schemes converge to optimal solutions.
\begin{table}[b]
  \centering
  \caption{Mean accuracies (\%± standard deviation) for large clients on Stanford40}
  \label{tab:scalability}
  \begin{tabular}{ccc}  
    \toprule
    {} & \multicolumn{2}{c}{Stanford40}  \\
    \cmidrule{2-3} 
    
        & $N=50$ &  $N=100$   \\
    \midrule
    SOLO &  $45.04 \pm 8.99$ &  $42.24 \pm 6.88$ \\
    \midrule
    FedAvg \cite{McMahan2016} &  $77.96 \pm 7.73$ &  $75.26 \pm 10.14$ \\
    \midrule
    SCAFFOLD \cite{Karimireddy2020} &  $81.63 \pm 6.04$ &  $82.95 \pm 8.41$  \\
    \midrule
    FedProx \cite{FedProx2020} &  $78.18 \pm 8.31$ &  $75.31 \pm 12.00$ \\
    \midrule
    MOON \cite{MOON2021} &  $78.60 \pm 8.19$ &  $76.29 \pm 7.97$ \\
    \midrule
    FedDyn \cite{Durmus2021} &  $82.28 \pm 10.85$ & $82.43 \pm 13.06$ \\
    \midrule
    FedHKD \cite{chen2023best} &  $78.91 \pm 10.55$ &  $76.15 \pm 8.45$ \\
    \midrule
    CDFL &  $\mathbf{83.01 \pm 4.49}$ &  $\mathbf{83.87 \pm 4.98}$ \\
    \bottomrule
  \end{tabular}
\end{table}
The performance of various schemes on a large number of clients for the Stanford40 dataset is presented in Table \ref{tab:scalability}. As expected, the SOLO approach exhibits a significant drop in performance when the number of clients increases. Among the baseline methods we considered, FedProx~\cite{FedProx2020}, MOON~\cite{MOON2021}, and FedHKD~\cite{chen2023best} perform similarly to FedAvg~\cite{McMahan2016}, struggling to handle a large number of clients effectively. In contrast, SCAFFOLD~\cite{Karimireddy2020} and FedDyn~\cite{Durmus2021} show improved performance compared to the other baselines in large-scale settings.
As shown in Table \ref{tab:scalability}, CDFL consistently outperforms the other baselines in the case of large-scale FL setups. The higher mean performance of CDFL implies its ability to effectively manage a large number of clients. Furthermore, its lower standard deviation compared to other baselines demonstrates its consistent performance across all the individual clients.

\begin{table*}[!h]
  \centering
  \caption{Mean accuracies (\%± standard deviation) for different levels of heterogeneity (\#clients=10)}
  \label{tab:heterogeneity}
  \begin{tabular}{ccccccccc}  
    \toprule
    {} & \multicolumn{2}{c}{Stanford40} & {} & \multicolumn{2}{c}{PPMI} & {} & \multicolumn{2}{c}{VOC2012} \\
    \cmidrule{2-3} 
    \cmidrule{5-6}
    \cmidrule{8-9}
        & $\alpha=0.1$ &  $\alpha=10.0$ & & $\alpha=0.1$  & $\alpha=10.0$ & & $\alpha=0.1$ &  $\alpha=10.0$   \\
    \midrule
    SOLO &  $59.82 \pm 14.93$ &  $71.52 \pm 2.26$ &   & $58.57 \pm 20.62$ &  $74.20 \pm 2.30$ &   & $61.24 \pm 25.85$ &  $57.76 \pm 12.77$ \\
    \midrule
    FedAvg \cite{McMahan2016} &  $78.31 \pm 6.45$ & $84.25 \pm 1.73$ &   & $72.74 \pm 13.68$ & $86.91 \pm 5.96$ &   & $73.73 \pm 21.43$ & $73.45 \pm 13.36$ \\
    \midrule
    SCAFFOLD \cite{Karimireddy2020} &  $81.31 \pm 4.52$ & $83.66 \pm 1.82$ &   & $76.39 \pm 12.20$ & $87.58 \pm 3.09$ &   & $69.71 \pm 13.32$ &  $68.08 \pm 14.02$ \\
    \midrule
    FedProx \cite{FedProx2020} &  $80.11 \pm 6.52$ & $84.08 \pm 2.17$ &   & $74.09 \pm 15.06$ & $87.57 \pm 4.48$ &   & $75.36 \pm 20.46$ & $\mathbf{78.61 \pm 3.45}$ \\
    \midrule
    MOON \cite{MOON2021} &  $78.49 \pm 6.02$ & $85.06 \pm 2.32$ &   & $73.51 \pm 17.03$ &  $87.35 \pm 5.94$ &   & $74.89 \pm 18.34$ & $74.73 \pm 9.82$ \\
    \midrule
    FedDyn \cite{Durmus2021} &  $83.93 \pm 4.83$ & $84.25 \pm 1.76$ &   & $80.21 \pm 10.62$ & $87.89 \pm 3.78$ &   & $74.74 \pm 13.15$ &  $75.33 \pm 5.82$ \\
    \midrule
    FedHKD \cite{chen2023best} &  $80.22 \pm 4.16$ & $83.57 \pm 3.09$ &   & $77.09 \pm 12.79$ & $86.15 \pm 10.74$ &   & $75.59 \pm 15.30$ & $74.1 \pm 13.01$ \\
    \midrule
    CDFL &  $\mathbf{84.16 \pm 4.94}$ & $\mathbf{85.15 \pm 1.19}$ &   & $\mathbf{82.74 \pm 11.29}$ & $\mathbf{90.08 \pm 2.02}$ &   & $\mathbf{76.09 \pm 13.62}$ & $75.09 \pm 6.51$ \\
    \bottomrule
  \end{tabular}
\end{table*}
\subsection{Heterogeneity}
To assess the performance of our CDFL under different levels of data heterogeneity, we conduct experiments using two different values of $\alpha$ across all datasets. Specifically, we tested $\alpha=0.1$ to represent a higher level of heterogeneity as well as $\alpha=10$ to consider less heterogeneous data. The results are presented in Table \ref{tab:heterogeneity}.

CDFL consistently outperforms other baselines when data heterogeneity is high ($\alpha=0.1$), demonstrating its robustness in handling challenging, heterogeneous situations. In this scenario, CDFL achieves a performance improvement of up to $\sim6\%$, 10\%, and 6\% compared to other baselines for Stanford40, PPMI, and VOC2012, respectively.

For $\alpha=10.0$, CDFL achieves the highest accuracy for the Stanford40 and PPMI datasets, while FedProx leads to the best accuracy for VOC2012. Generally, the performance of all schemes is quite similar under homogeneous conditions ($\alpha=10.0$), which demonstrates the ability of FL systems to obtain desired performance in homogeneous data settings.

\subsection{Efficiency and Convergence Speed}
we compare CDFL with other baselines in terms of the number of communication rounds required to achieve a specific target accuracy as well as communication overhead in each communication round. 
Table \ref{tab:efficiency} provides an overview of the communication rounds needed to reach the test target accuracy for various numbers of clients across all datasets. In all settings, we continue training until we reach the target test accuracy or reach a maximum of 50 communication rounds.

As illustrated in Table \ref{tab:efficiency}, CDFL requires the fewest communication rounds in all settings and demonstrates a remarkable increase in convergence speed up to 10 times faster compared to other baselines. %In the large-scale Stanford40 setting (with 50 and 100 clients), which is similar to realistic scenarios, our proposed algorithm converges significantly faster than other baselines. 
For the VOC2012 dataset with 10 clients, both SCAFFOLD~\cite{Karimireddy2020} and FedHKD~\cite{chen2023best} failed to reach the target accuracy after 50 communication rounds.

Each communication round involves two types of communication overhead: uplink and downlink. Uplink includes the transmission of local models to the server, while downlink corresponds to broadcasting the updated global model to the clients. As the downlink overhead is considerably lower than the uplink, our analysis focuses on the uplink overhead. 
In this study, we define the number of tansmitted parameters as communication overhead metric.

%In this study, we consider the amounts of parameters that are exchanged between the server and the clients as the communication overhead
\begin{table*}[t]
  \centering
  \caption{Comparison of the number of communication rounds for different FL schemes to achieve the target accuracy }
  \label{tab:efficiency}
  \resizebox{\textwidth}{!}{\begin{tabular}{cccccccccccccc}  
    \toprule
    {} & \multicolumn{5}{c}{Stanford40} & {} & \multicolumn{3}{c}{PPMI} & {} & \multicolumn{3}{c}{VOC2012} \\
    \cmidrule{2-6} 
    \cmidrule{8-10}
    \cmidrule{12-14}
        & $N=10$ & $N=20$ & $N=30$ & $N=50$ & $N=100$ & & $N=10$ & $N=20$ & $N=30$ & & $N=10$ & $N=20$ & $N=30$  \\
    \cmidrule{2-6} 
    \cmidrule{8-10}
    \cmidrule{12-14}
    Target & 85\% & 80\% & 80\% & 80\% & 80\% & & 85\% & 85\% & 80\% & & 75\% & 70\% & 70\%  \\
    \midrule
    
    FedAvg \cite{McMahan2016} & 14 ($0.6 \times$) & 16 ($0.5 \times$) & 24 ($0.4 \times$)& 35 ($0.3 \times$)& 50 ($0.1 \times$)& & 9 ($0.6 \times$)& 34 ($0.3 \times$)& 37 ($0.2 \times$)& & 34 ($0.3 \times$)& 44 ($0.2 \times$)& 38 ($0.2 \times$) \\
    \midrule
    SCAFFOLD \cite{Karimireddy2020} & 22 ($0.4 \times$) & 12 ($0.7 \times$)& 13 ($0.7 \times$)& 16 ($0.6 \times$)& 17 ($0.4 \times$)& & 19 ($0.3 \times$)& 31 ($0.3 \times$)& 12 ($0.7 \times$)& & - & 10 ($0.8 \times$)& 11 ($0.8 \times$) \\
    \midrule
    FedProx \cite{FedProx2020} & 10 ($0.8 \times$)& 15 ($0.5 \times$)& 24 ($0.4 \times$)& 35 ($0.3 \times$)& 50 ($0.1 \times$)& & 11 ($0.5 \times$)& 35 ($0.3 \times$)& 38 ($0.2 \times$)& & 18 ($0.6 \times$)& 10 ($0.8 \times$)& 25 ($0.4 \times$) \\
    \midrule
    MOON \cite{MOON2021} & 11 ($0.7 \times$)& 13 ($0.6 \times$)& 19 ($0.5 \times$)& 35 ($0.3 \times$)& 46 ($0.1 \times$)& & 14 ($0.4 \times$)& 34 ($0.3 \times$)& 30 ($0.3 \times$)& & 29 ($0.4 \times$)& 31 ($0.3 \times$)& 31 ($0.3 \times$) \\
    \midrule
    FedDyn \cite{Durmus2021} & 14 ($0.6 \times$)& \textbf{8 ($\mathbf{1\times}$)}& 10 ($0.9 \times$)& 17 ($0.6 \times$)& 18 ($0.3 \times$)& & 14 ($0.4 \times$)& 26 ($0.3 \times$)& 12 ($0.7 \times$)& & 30 ($0.4 \times$)& 15 ($0.5 \times$)& \textbf{9 ($\mathbf{1\times}$)} \\
    \midrule
    FedHKD \cite{chen2023best} & \textbf{8 ($\mathbf{1\times}$)}& 11 ($0.7 \times$)& 15 ($0.6 \times$)& 29 ($0.3 \times$)& 46 ($0.1 \times$)& & 6 ($0.8 \times$)& 36 ($0.3 \times$)& 32 ($0.3 \times$)& & - & 30 ($0.3 \times$)& 31 ($0.3 \times$) \\
    \midrule
    CDFL & \textbf{8 ($\mathbf{1\times}$)} & \textbf{8 ($\mathbf{1\times}$)} & \textbf{9 ($\mathbf{1\times}$)} & \textbf{10 ($\mathbf{1\times}$)} & \textbf{6 ($\mathbf{1\times}$)} & & \textbf{5 ($\mathbf{1\times}$)} & \textbf{9 ($\mathbf{1\times}$)}& \textbf{8 ($\mathbf{1\times}$)}& & \textbf{11 ($\mathbf{1\times}$)}& \textbf{8 ($\mathbf{1\times}$)}& \textbf{9 ($\mathbf{1\times}$)} \\
    \bottomrule
  \end{tabular}}
\end{table*}

In Table \ref{tab:overhead}, we summarize the communication overhead of various FL schemes. Since the uplink communication overhead of FedProx~\cite{FedProx2020}, MOON~\cite{MOON2021}, and FedDyn~\cite{Durmus2021} matches that of FedAvg~\cite{McMahan2016}, we only report the uplink overhead of FedAvg~\cite{McMahan2016} in Table \ref{tab:overhead}. In this table, $p$ represents the ratio of connected clients to the total number of clients, and $|\theta_i|$ is the number of parameters in each client's model. In the case of SCAFFOLD~\cite{Karimireddy2020}, $c_i$ is the control variate whose size matches the client's model ($|c_i|=|\theta_i|$). For the FedHKD~\cite{chen2023best} scheme, hyperknowledge vectors are transmitted to the server alongside the clients' models, and $|\mathbf{v}|$ represents the size of the penultimate layer in the model. In most cases, the overhead arising from hyperknowledge vectors is negligible compared to the model transmission overhead. In CDFL, the pixelized images are transmitted beside the clients' models but for a smaller number of clients ($|\mathcal{C}|$).

 We evaluate Stanford40, PPMI, and VOC2012 datasets with 100, 30, and 30 clients, respectively. We select the largest number of clients for each dataset, as the communication overhead would be more challenging with a larger number of clients. The overhead of all schemes is calculated in Table \ref{tab:overhead}. For CDFL, we assumed the size of images to be $|I|=224\times224\times3$, and this is the same as the input size of ResNet50. However, there exists an opportunity to compress or downsample the images before transmission to the server, which can save more bandwidth. Since ResNet50 is utilized in our experiments, the total number of parameters is 24.1 M.%$|\theta_i|$ is approximately $24.10\times 10^6$.

As shown in Table \ref{tab:heterogeneity}, SCAFFOLD has the highest overhead among all schemes, while CDFL provides the lowest overhead due to its client selection. We achieved a reduction in overhead by up to $\sim$ 64\% compared to other baselines for all datasets.%We achieved a reduction in overhead by up to 63\%, 63\%, and 64\% compared to other schemes for Stanford40, PPMI, and VOC2012, respectively.

\begin{table}[t!]
  \centering
  \caption{Communication overhead ($\times 10^8$) in each round of different FL schemes }
  \label{tab:overhead}
  \resizebox{\columnwidth}{!}{\begin{tabular}{ccccc}  
    \toprule
    {} & Uplink & Stanford40 & PPMI & VOC2012  \\

    \midrule
    FedAvg \cite{McMahan2016} & $pN |\theta_i|$ & $19.3 $ & $5.78$ & $5.78$ \\
    \midrule
    SCAFFOLD \cite{Karimireddy2020} & $pN (|\theta_i|+ |c_i|)$ & $38.6$ & $11.6$ & $11.6$ \\
    \midrule
    %FedProx \cite{FedProx2020} & & $1.93 \times 10^9$ & $5.78 \times 10^8$ & $5.78 \times 10^8$ \\
    %\midrule
    %MOON \cite{MOON2021} & & $1.93 \times 10^9$ & $5.78 \times 10^8$ & $5.78 \times 10^8$ \\
    %\midrule
    %FedDyn \cite{Durmus2021} & & $1.93 \times 10^9$ & $5.78 \times 10^8$ & $5.78 \times 10^8$ \\
    %\midrule
    FedHKD \cite{chen2023best} & $pN (|\theta_i|+|C|(|\mathbf{v}|+|C|))$ & $19.3$ & $5.78$ & $5.78$ \\
    \midrule
    CDFL & $|\mathcal{C}| (|\theta_i|+|\mathcal{S}_i| |I|)$ & $\mathbf{14.3}$ & $\mathbf{4.32}$ & $\mathbf{4.19}$ \\
    \bottomrule
  \end{tabular}}
\end{table}

\subsection{Analysis of CDFL}
In this section, we study the effects of key modules and hyper-parameters in CDFL. The experiments are mainly conducted on the PPMI datasets with 10 clients and $\alpha=1.0$. 
\begin{itemize}
    \item \textit{Imapact of Contrastive Learning \& Deep Clustering Modules}
    
    We investigate the effect of the contrastive learning and deep clustering modules by excluding them from CDFL, as shown in Table \ref{tab:removal}. The contrastive learning module is eliminated by setting $\lambda_1=\lambda_2=0$ in \textit{local} model loss function. Therefore, this test only relies on the training of pixelized images. In the second test, the deep clustering term is omitted from the \textit{local} model loss function by setting $\lambda_3$ to zero, and random pixelized images are chosen for transmission to the server. As presented in Table \ref{tab:removal}, removing the contrastive learning module from CDFL results in a 1.27\% decrease in performance. In addition, eliminating the deep clustering modules and transmitting random pixelized images to the server leads to a 3.9\% reduction in performance, highlighting the effectiveness of our image selection method.

    \begin{table}[h!]
    \centering
    \caption{Impact of contrastive learning and deep clustering modules of CDFL}
    \label{tab:removal}
    \begin{tabular}{cc}  
    \toprule
     Removing contrastive learning & Removing deep clustering    \\
    \midrule
    $89.47 \pm 5.368$ ($1.27 \% \downarrow$) & $86.84 \pm 4.30$ ($3.9\% \downarrow$)\\ %\hline
    %VOC2012 &$74.58 \pm 2.34$ &  $77.82 \pm 2.49$ & $76.11 \pm 1.53$\\
    \bottomrule
    \end{tabular}
    \end{table}
    \item \textit{Impact of $\lambda_1$ and $\lambda_2$ of Model Contrastive Learning}
    
    We investigate the effect of various values of $\lambda_1$ and $\lambda_2$ on the performance of CDFL. Test performance results for different values of $\lambda_1$ and $\lambda_2$ are presented in Table \ref{tab:lambda12}. An increase in $\lambda_1$ results in a more similar representation between the local model on both original and pixelized images. Further, larger values of $\lambda_2$ lead to a more similar representation between \textit{personalized} and \textit{local} models on original images. 
    \begin{table}[h!]
    \centering
    \caption{Impact of $\lambda_1$ and $\lambda_2$ on the performance of CDFL}
    \label{tab:lambda12}
    \begin{tabular}{ccc}  
    \toprule
     $\lambda_1=\lambda_2=0.01$ & $\lambda_1=\lambda_2=0.1$ &  $\lambda_1=\lambda_2=1.0$   \\
    \midrule
    $89.63 \pm 4.31$ &  $90.74 \pm 2.98$ & $88.68 \pm 3.23$\\ %\hline
    %VOC2012 &$74.58 \pm 2.34$ &  $77.82 \pm 2.49$ & $76.11 \pm 1.53$\\
    \bottomrule
    \end{tabular}
    \end{table}

    \item \textit{Impact of $\lambda_3$ in Deep Clustering}

    Deep clustering is a crucial element in CDFL as it helps select representative images for the clients. The impact of $\lambda_3$ in deep clustering loss is investigated for three different values of $\{0.01,0.05,0.1 \}$, as summarized in Table \ref{tab:lambda3}. Smaller values of $\lambda_3$ result in poor clustering and image selection, while larger values of $\lambda_3$ fade the impact of contrastive learning and knowledge distillation terms in the local loss function, causing the local model to diverge from the \textit{personalized} model.

    \begin{table}[h!]
    \centering
    \caption{Impact of $\lambda_3$ on the performance of CDFL}
    \label{tab:lambda3}
    \begin{tabular}{ccc}  
    \toprule
    $\lambda_3=0.01$ & $\lambda_3=0.0.5$ &  $\lambda_3=0.1$   \\
    \midrule
    $90.11 \pm 2.30$ &  $90.74 \pm 2.98$ & $90.19 \pm 3.69$\\ %\hline
    %VOC2012 & $74.83 \pm 3.07$ &  $77.82 \pm 2.49$ & $75.41 \pm 2.62$\\
    \bottomrule
    \end{tabular}
    \end{table}

    \item \textit{Impact of Number of Selected Images}
    \begin{table}[b]
    \centering
    \caption{Impact of the number of selected images on the performance of CDFL}
    \label{tab:m}
    \begin{tabular}{cccc}  
    \toprule 
    $m=1$ & $m=3$ & $m=5$ &  $m=7$   \\
    \midrule
    $88.13 \pm 5.39$ & $89.32 \pm 4.39$ & $88.77 \pm 5.26$ &  $90.74 \pm 2.98$ \\ %\hline
    %VOC2012 & $88.13 \pm 5.39$ & $73.45 \pm 3.21$ & $75.38 \pm 4.04$ &  $77.82 \pm 2.49$ \\
    \bottomrule
    \end{tabular}
    \end{table}
    In CDFL, we select the $m$ nearest neighbors of each centroid for transmission to the server. A very small $m$ selection leads to a non-representative image set, while a large $m$ increases communication overhead. The performance of CDFL for four different values of $m$ is shown in Table \ref{tab:m}. As shown in Table \ref{tab:m}, smaller $m$ values, while slightly sacrificing performance, reduce bandwidth usage and communication overhead. This shows that CDFL can be effective even with fewer selected images.

    \item \textit{Impact of Number of Selected Clients}
    
    We also study the impact of the number of selected clients in CDFL. We repeat the experiments when only 3, 5, and 7 clients share their weights and images with the central server, and the results are summarized in Table \ref{tab:C}. As expected, when fewer clients contribute to the server aggregation, the performance drops, however, the performance drop is negligible. This demonstrates that CDFL works well even if only 30\% of clients are involved in server aggregation resulting in a significant decrease in communication overhead.

    \begin{table}[h!]
    \centering
    \caption{Impact of the number of selected clients on the performance of CDFL }
    \label{tab:C}
    \begin{tabular}{ccc}  
    \toprule 
    $|\mathcal{C}|=3$ & $|\mathcal{C}|=5$ &  $|\mathcal{C}|=7$   \\
    \midrule
    $89.48 \pm 4.44$ &  $90.74 \pm 2.98$ & $91.02 \pm 2.65$ \\ 
    %VOC2012 & $69.54 \pm 2.58$ &  $77.82 \pm 2.49$ & $91.02 \pm 2.65$ \\
    \bottomrule
    \end{tabular}
    \end{table}

    %\item \textit{Impact of Number of Local Epochs}
    
    %The impact of different numbers of local epochs is also studied in CDFL, as reported in Table \ref{tab:e}. Setting the number of local epochs to 1 results in minimal local updates, leading to slow training and lower performance. On the other hand, selecting a large number of local epochs moves the clients toward their local optima, hence deviating them from the global objective, which is not desirable.
    %\begin{table}[h!]
    %\centering
    %\caption{Impact of different numbers of local epochs on the performance of CDFL}
    %\label{tab:e}
    %\begin{tabular}{ccc}  
    %\toprule 
    %$e=1$ & $e=3$ &  $e=5$   \\
    %\midrule
    %$88.94 \pm 1.52$ & $89.78 \pm 4.28$ & $90.74 \pm 2.98$  \\ %\hline
    %VOC2012 & $77.82 \pm 2.75$ & $76.47 \pm 4.11$ &$77.82 \pm 2.49$   \\
    %\bottomrule
    %\end{tabular}
    %\end{table}
\end{itemize}

\section{Conclusion}
\label{section:conclusion}

This paper introduces CDFL as an efficient FL framework for recognizing human activities. It is designed specifically to address the prevalent challenges of data heterogeneity, communication overhead, and naive aggregation. CDFL introduces a privacy-preserved subset of samples to the central server to enhance the performance of the system in a secure manner. %without harming clients' security. 
To optimize the selection of representative data samples for each activity, we integrated methods of contrastive learning, knowledge distillation, and deep clustering. %Using this enhanced representation,  the framework effectively overcomes the data heterogeneity, outperforming state-of-the-art methods. 
Furthermore, our new client selection method offers the dual advantage of reducing communication overheads and enhancing performance. Extensive experiments on three well-known image-based HAR datasets highlight the efficacy of CDFL in both classification accuracy and communication efficiency. Overall, a performance increase of up to 10\%, a bandwidth usage reduction of 64\% as well as a faster convergence rate of up to 10 times in comparison to the state-of-the-art schemes is achieved.  
%This paper introduces a unique Federated Learning (FL) framework for Human Activity Recognition (HAR) designed specifically to address the prevalent challenges of data heterogeneity and data distribution across clients in standard FL architectures. Key to our approach is the incorporation of a privacy-preserved data subset to the central server, ensuring both enhanced security and data integrity. To optimize the selection of representative data samples for each activity, we integrated methods of contrastive learning, knowledge distillation, and deep clustering. Using this enhanced representation,  the framework effectively overcomes the data heterogeneity, outperforming state-of-the-art methods. Furthermore, the developed client selection methodology offers the dual advantage of reducing communication overheads, thereby making the system more efficient and scalable. Rigorous experiments on three renowned HAR datasets highlight the efficacy of the proposed framework in both classification accuracy and communication efficiency. Overall, a performance increase of 10\%, a bandwidth usage reduction of 28\% as well as a faster convergence rate of 10 times in comparison to the SOTA is achieved.  

In the future, several directions of the proposed work will be investigated. First, we want to study the influence of degraded image quality on the efficacy of our system. This assessment is pivotal given the real-world scenarios where image quality can vary depending on the sensor and external factors. Furthermore, we will design a robust and efficient FL framework for the multimodal HAR task that combines signals from various modalities, including visual, audio, inertial, and biological signals.

%\bibliographystyle{IEEEtran}
%\bibliography{IEEEabrv,./refs}

\begin{thebibliography}{10}
\providecommand{\url}[1]{#1}
\csname url@samestyle\endcsname
\providecommand{\newblock}{\relax}
\providecommand{\bibinfo}[2]{#2}
\providecommand{\BIBentrySTDinterwordspacing}{\spaceskip=0pt\relax}
\providecommand{\BIBentryALTinterwordstretchfactor}{4}
\providecommand{\BIBentryALTinterwordspacing}{\spaceskip=\fontdimen2\font plus
\BIBentryALTinterwordstretchfactor\fontdimen3\font minus \fontdimen4\font\relax}
\providecommand{\BIBforeignlanguage}[2]{{%
\expandafter\ifx\csname l@#1\endcsname\relax
\typeout{** WARNING: IEEEtran.bst: No hyphenation pattern has been}%
\typeout{** loaded for the language `#1'. Using the pattern for}%
\typeout{** the default language instead.}%
\else
\language=\csname l@#1\endcsname
\fi
#2}}
\providecommand{\BIBdecl}{\relax}
\BIBdecl

\bibitem{guo2018multisensor}
M.~Guo, Z.~Wang, N.~Yang, Z.~Li, and T.~An, ``A multisensor multiclassifier hierarchical fusion model based on entropy weight for human activity recognition using wearable inertial sensors,'' \emph{IEEE Transactions on Human-Machine Systems}, vol.~49, no.~1, pp. 105--111, 2018.

\bibitem{gaglio2014human}
S.~Gaglio, G.~L. Re, and M.~Morana, ``Human activity recognition process using 3-d posture data,'' \emph{IEEE Transactions on Human-Machine Systems}, vol.~45, no.~5, pp. 586--597, 2014.

\bibitem{kamel2018deep}
A.~Kamel, B.~Sheng, P.~Yang, P.~Li, R.~Shen, and D.~D. Feng, ``Deep convolutional neural networks for human action recognition using depth maps and postures,'' \emph{IEEE Transactions on Systems, Man, and Cybernetics: Systems}, vol.~49, no.~9, pp. 1806--1819, 2018.

\bibitem{dang2020sensor}
L.~M. Dang, K.~Min, H.~Wang, M.~J. Piran, C.~H. Lee, and H.~Moon, ``Sensor-based and vision-based human activity recognition: A comprehensive survey,'' \emph{Pattern Recognition}, vol. 108, p. 107561, 2020.

\bibitem{esmaeilzehi2024harwe}
A.~Esmaeilzehi, E.~Khazaei, K.~Wang, N.~K. Kalsi, P.~C. Ng, H.~Liu, Y.~Yu, D.~Hatzinakos, and K.~Plataniotis, ``Harwe: A multi-modal large-scale dataset for context-aware human activity recognition in smart working environments,'' \emph{Pattern Recognition Letters}, 2024.

\bibitem{zhou2020deep}
X.~Zhou, W.~Liang, I.~Kevin, K.~Wang, H.~Wang, L.~T. Yang, and Q.~Jin, ``Deep-learning-enhanced human activity recognition for internet of healthcare things,'' \emph{IEEE Internet of Things Journal}, vol.~7, no.~7, pp. 6429--6438, 2020.

\bibitem{moniruzzaman2021human}
M.~Moniruzzaman, Z.~Yin, Z.~He, R.~Qin, and M.~C. Leu, ``Human action recognition by discriminative feature pooling and video segment attention model,'' \emph{IEEE Transactions on Multimedia}, vol.~24, pp. 689--701, 2021.

\bibitem{qi2020smartphone}
W.~Qi, H.~Su, and A.~Aliverti, ``A smartphone-based adaptive recognition and real-time monitoring system for human activities,'' \emph{IEEE Transactions on Human-Machine Systems}, vol.~50, no.~5, pp. 414--423, 2020.

\bibitem{McMahan2016}
\BIBentryALTinterwordspacing
H.~B. McMahan, E.~Moore, D.~Ramage, S.~Hampson, and B.~A. y~Arcas, ``Communication-efficient learning of deep networks from decentralized data,'' in \emph{International Conference on Artificial Intelligence and Statistics}, 2016. [Online]. Available: \url{https://api.semanticscholar.org/CorpusID:14955348}
\BIBentrySTDinterwordspacing

\bibitem{FedProx2020}
T.~Li, A.~Sahu, M.~Zaheer, M.~Sanjabi, A.~Talwalkar, and V.~Smith, ``Federated optimization in heterogeneous networks,'' in \emph{Third Conference on Machine Learning and Systems (MLSys)}, vol.~2, 2020, p. 429–450.

\bibitem{APFL2020}
\BIBentryALTinterwordspacing
Y.~Deng, M.~M. Kamani, and M.~Mahdavi, ``Adaptive personalized federated learning,'' \emph{CoRR}, vol. abs/2003.13461, 2020. [Online]. Available: \url{https://arxiv.org/abs/2003.13461}
\BIBentrySTDinterwordspacing

\bibitem{wang2019}
K.~Wang, R.~Mathews, C.~Kiddon, H.~Eichner, F.~Beaufays, and D.~Ramage, ``Federated evaluation of on-device personalization,'' 2019.

\bibitem{arivazhagan2019}
M.~G. Arivazhagan, V.~Aggarwal, A.~K. Singh, and S.~Choudhary, ``Federated learning with personalization layers,'' 2019.

\bibitem{jiang2023}
Y.~Jiang, J.~Konečný, K.~Rush, and S.~Kannan, ``Improving federated learning personalization via model agnostic meta learning,'' 2023.

\bibitem{Huang2020}
\BIBentryALTinterwordspacing
Y.~Huang, L.~Chu, Z.~Zhou, L.~Wang, J.~Liu, J.~Pei, and Y.~Zhang, ``Personalized cross-silo federated learning on non-iid data,'' in \emph{AAAI Conference on Artificial Intelligence}, 2020. [Online]. Available: \url{https://api.semanticscholar.org/CorpusID:227311284}
\BIBentrySTDinterwordspacing

\bibitem{Dinh2020}
C.~T. Dinh, N.~H. Tran, and T.~D. Nguyen, ``Personalized federated learning with moreau envelopes,'' in \emph{Proceedings of the 34th International Conference on Neural Information Processing Systems}, ser. NIPS'20.\hskip 1em plus 0.5em minus 0.4em\relax Red Hook, NY, USA: Curran Associates Inc., 2020.

\bibitem{Fallah2020}
A.~Fallah, A.~Mokhtari, and A.~Ozdaglar, ``Personalized federated learning with theoretical guarantees: A model-agnostic meta-learning approach,'' in \emph{Proceedings of the 34th International Conference on Neural Information Processing Systems}, ser. NIPS'20.\hskip 1em plus 0.5em minus 0.4em\relax Red Hook, NY, USA: Curran Associates Inc., 2020.

\bibitem{cFedFCN2023}
X.-X. Wei and H.~Huang, ``Edge devices clustering for federated visual classification: A feature norm based framework,'' \emph{IEEE Transactions on Image Processing}, vol.~32, pp. 995--1010, 2023.

\bibitem{Sattler2021}
F.~Sattler, K.-R. Müller, and W.~Samek, ``Clustered federated learning: Model-agnostic distributed multitask optimization under privacy constraints,'' \emph{IEEE Transactions on Neural Networks and Learning Systems}, vol.~32, no.~8, pp. 3710--3722, 2021.

\bibitem{Ghosh2022}
A.~Ghosh, J.~Chung, D.~Yin, and K.~Ramchandran, ``An efficient framework for clustered federated learning,'' \emph{IEEE Transactions on Information Theory}, vol.~68, no.~12, pp. 8076--8091, 2022.

\bibitem{Chen2020}
C.~Chen, Z.~Chen, Y.~Zhou, and B.~Kailkhura, ``Fedcluster: Boosting the convergence of federated learning via cluster-cycling,'' in \emph{2020 IEEE International Conference on Big Data (Big Data)}, 2020, pp. 5017--5026.

\bibitem{Briggs2020}
C.~Briggs, Z.~Fan, and P.~Andras, ``Federated learning with hierarchical clustering of local updates to improve training on non-iid data,'' in \emph{2020 International Joint Conference on Neural Networks (IJCNN)}, 2020, pp. 1--9.

\bibitem{Duan2022}
M.~Duan, D.~Liu, X.~Ji, Y.~Wu, L.~Liang, X.~Chen, Y.~Tan, and A.~Ren, ``Flexible clustered federated learning for client-level data distribution shift,'' \emph{IEEE Transactions on Parallel and Distributed Systems}, vol.~33, no.~11, pp. 2661--2674, 2022.

\bibitem{Ouyang2021}
\BIBentryALTinterwordspacing
X.~Ouyang, Z.~Xie, J.~Zhou, J.~Huang, and G.~Xing, ``Clusterfl: A similarity-aware federated learning system for human activity recognition,'' in \emph{Proceedings of the 19th Annual International Conference on Mobile Systems, Applications, and Services}, ser. MobiSys '21.\hskip 1em plus 0.5em minus 0.4em\relax New York, NY, USA: Association for Computing Machinery, 2021, p. 54–66. [Online]. Available: \url{https://doi.org/10.1145/3458864.3467681}
\BIBentrySTDinterwordspacing

\bibitem{Eiffel2022}
A.~Sultana, M.~M. Haque, L.~Chen, F.~Xu, and X.~Yuan, ``Eiffel: Efficient and fair scheduling in adaptive federated learning,'' \emph{IEEE Transactions on Parallel and Distributed Systems}, vol.~33, no.~12, pp. 4282--4294, 2022.

\bibitem{Chen2023}
D.~Chen, J.~Hu, V.~Junkai~Tan, X.~Wei, and E.~Wu, ``Elastic aggregation for federated optimization,'' in \emph{2023 IEEE/CVF Conference on Computer Vision and Pattern Recognition (CVPR)}, 2023.

\bibitem{MOON2021}
Q.~Li, B.~He, and D.~Song, ``Model-contrastive federated learning,'' in \emph{2021 IEEE/CVF Conference on Computer Vision and Pattern Recognition (CVPR)}, 2021, pp. 10\,708--10\,717.

\bibitem{FedCluster2020}
C.~Chen, Z.~Chen, Y.~Zhou, and B.~Kailkhura, ``Fedcluster: Boosting the convergence of federated learning via cluster-cycling,'' in \emph{2020 IEEE International Conference on Big Data (Big Data)}, 2020, pp. 5017--5026.

\bibitem{FedGroup2021}
M.~Duan, D.~Liu, X.~Ji, R.~Liu, L.~Liang, X.~Chen, and Y.~Tan, ``Fedgroup: Efficient federated learning via decomposed similarity-based clustering,'' in \emph{2021 IEEE Intl Conf on Parallel \& Distributed Processing with Applications, Big Data \& Cloud Computing, Sustainable Computing \& Communications, Social Computing \& Networking (ISPA/BDCloud/SocialCom/SustainCom)}, 2021, pp. 228--237.

\bibitem{albaseer2023data}
A.~Albaseer, M.~Abdallah, A.~Al-Fuqaha, and A.~Erbad, ``Data-driven participant selection and bandwidth allocation for heterogeneous federated edge learning,'' \emph{IEEE Transactions on Systems, Man, and Cybernetics: Systems}, 2023.

\bibitem{Chen2021}
\BIBentryALTinterwordspacing
M.~Chen, N.~Shlezinger, H.~V. Poor, Y.~C. Eldar, and S.~Cui, ``Communication-efficient federated learning,'' \emph{Proceedings of the National Academy of Sciences}, vol. 118, no.~17, p. e2024789118, 2021. [Online]. Available: \url{https://www.pnas.org/doi/abs/10.1073/pnas.2024789118}
\BIBentrySTDinterwordspacing

\bibitem{Chen2020ASWT_FedAvg}
Y.~Chen, X.~Sun, and Y.~Jin, ``Communication-efficient federated deep learning with layerwise asynchronous model update and temporally weighted aggregation,'' \emph{IEEE Transactions on Neural Networks and Learning Systems}, vol.~31, no.~10, pp. 4229--4238, 2020.

\bibitem{Rothchild2020}
\BIBentryALTinterwordspacing
D.~Rothchild, A.~Panda, E.~Ullah, N.~Ivkin, I.~Stoica, V.~Braverman, J.~Gonzalez, and R.~Arora, ``{F}etch{SGD}: Communication-efficient federated learning with sketching,'' in \emph{Proceedings of the 37th International Conference on Machine Learning}, ser. Proceedings of Machine Learning Research, H.~D. III and A.~Singh, Eds., vol. 119.\hskip 1em plus 0.5em minus 0.4em\relax PMLR, 13--18 Jul 2020, pp. 8253--8265. [Online]. Available: \url{https://proceedings.mlr.press/v119/rothchild20a.html}
\BIBentrySTDinterwordspacing

\bibitem{FedMA2020}
\BIBentryALTinterwordspacing
H.~Wang, M.~Yurochkin, Y.~Sun, D.~Papailiopoulos, and Y.~Khazaeni, ``Federated learning with matched averaging,'' in \emph{International Conference on Learning Representations}, 2020. [Online]. Available: \url{https://openreview.net/forum?id=BkluqlSFDS}
\BIBentrySTDinterwordspacing

\bibitem{EK2021}
S.~EK, F.~PORTET, P.~LALANDA, and G.~VEGA, ``A federated learning aggregation algorithm for pervasive computing: Evaluation and comparison,'' in \emph{2021 IEEE International Conference on Pervasive Computing and Communications (PerCom)}, 2021, pp. 1--10.

\bibitem{Zhou2022}
X.~Zhou, W.~Liang, J.~Ma, Z.~Yan, and K.~I.-K. Wang, ``2d federated learning for personalized human activity recognition in cyber-physical-social systems,'' \emph{IEEE Transactions on Network Science and Engineering}, vol.~9, no.~6, pp. 3934--3944, 2022.

\bibitem{Concone2022}
F.~Concone, C.~Ferdico, G.~L. Re, and M.~Morana, ``A federated learning approach for distributed human activity recognition,'' in \emph{2022 IEEE International Conference on Smart Computing (SMARTCOMP)}, 2022, pp. 269--274.

\bibitem{Xiao2021}
\BIBentryALTinterwordspacing
Z.~Xiao, X.~Xu, H.~Xing, F.~Song, X.~Wang, and B.~Zhao, ``A federated learning system with enhanced feature extraction for human activity recognition,'' \emph{Knowledge-Based Systems}, vol. 229, p. 107338, 2021. [Online]. Available: \url{https://www.sciencedirect.com/science/article/pii/S0950705121006006}
\BIBentrySTDinterwordspacing

\bibitem{Gad2023}
\BIBentryALTinterwordspacing
G.~Gad and Z.~Fadlullah, ``Federated learning via augmented knowledge distillation for heterogenous deep human activity recognition systems,'' \emph{Sensors}, vol.~23, no.~1, 2023. [Online]. Available: \url{https://www.mdpi.com/1424-8220/23/1/6}
\BIBentrySTDinterwordspacing

\bibitem{li2019fedmd}
D.~Li and J.~Wang, ``Fedmd: Heterogenous federated learning via model distillation,'' 2019.

\bibitem{Tu2021}
\BIBentryALTinterwordspacing
L.~Tu, X.~Ouyang, J.~Zhou, Y.~He, and G.~Xing, ``Feddl: Federated learning via dynamic layer sharing for human activity recognition,'' in \emph{Proceedings of the 19th ACM Conference on Embedded Networked Sensor Systems}, ser. SenSys '21.\hskip 1em plus 0.5em minus 0.4em\relax New York, NY, USA: Association for Computing Machinery, 2021, p. 15–28. [Online]. Available: \url{https://doi.org/10.1145/3485730.3485946}
\BIBentrySTDinterwordspacing

\bibitem{Bettini2022}
\BIBentryALTinterwordspacing
C.~Bettini, G.~Civitarese, and R.~Presotto, ``Semi-supervised and personalized federated activity recognition based on active learning and label propagation,'' \emph{Personal and Ubiquitous Computing}, vol.~26, no.~5, pp. 1281--1298, jun 2022. [Online]. Available: \url{https://doi.org/10.1007%2Fs00779-022-01688-8}
\BIBentrySTDinterwordspacing

\bibitem{zhao2021}
Y.~Zhao, H.~Liu, H.~Li, P.~Barnaghi, and H.~Haddadi, ``Semi-supervised federated learning for activity recognition,'' 2021.

\bibitem{Retinaface}
J.~Deng, J.~Guo, E.~Ververas, I.~Kotsia, and S.~Zafeiriou, ``Retinaface: Single-shot multi-level face localisation in the wild,'' in \emph{CVPR}, 2020.

\bibitem{Yao2011}
B.~Yao, X.~Jiang, A.~Khosla, A.~L. Lin, L.~Guibas, and L.~Fei-Fei, ``Human action recognition by learning bases of action attributes and parts,'' in \emph{2011 International Conference on Computer Vision}, 2011, pp. 1331--1338.

\bibitem{Hook2016}
K.~Sohn, ``Improved deep metric learning with multi-class n-pair loss objective,'' in \emph{Proceedings of the 30th International Conference on Neural Information Processing Systems}, ser. NIPS'16.\hskip 1em plus 0.5em minus 0.4em\relax Red Hook, NY, USA: Curran Associates Inc., 2016, p. 1857–1865.

\bibitem{Xu2023}
B.~Xu, W.~Xia, H.~Zhao, Y.~Zhu, X.~Sun, and T.~Q.~S. Quek, ``Clustered federated learning in internet of things: Convergence analysis and resource optimization,'' \emph{IEEE Internet of Things Journal}, pp. 1--1, 2023.

\bibitem{Shlezinger2020}
N.~Shlezinger, S.~Rini, and Y.~C. Eldar, ``The communication-aware clustered federated learning problem,'' in \emph{2020 IEEE International Symposium on Information Theory (ISIT)}, 2020, pp. 2610--2615.

\bibitem{Chengxi2022}
C.~Li, G.~Li, and P.~K. Varshney, ``Federated learning with soft clustering,'' \emph{IEEE Internet of Things Journal}, vol.~9, no.~10, pp. 7773--7782, 2022.

\bibitem{Kim2021}
Y.~Kim, E.~A. Hakim, J.~Haraldson, H.~Eriksson, J.~M.~B. da~Silva, and C.~Fischione, ``Dynamic clustering in federated learning,'' in \emph{ICC 2021 - IEEE International Conference on Communications}, 2021, pp. 1--6.

\bibitem{Xueyang2021}
\BIBentryALTinterwordspacing
X.~Tang, S.~Guo, and J.~Guo, ``Personalized federated learning with clustered generalization,'' \emph{CoRR}, vol. abs/2106.13044, 2021. [Online]. Available: \url{https://arxiv.org/abs/2106.13044}
\BIBentrySTDinterwordspacing

\bibitem{Jamali2022}
H.~Jamali-Rad, M.~Abdizadeh, and A.~Singh, ``Federated learning with taskonomy for non-iid data,'' \emph{IEEE Transactions on Neural Networks and Learning Systems}, pp. 1--12, 2022.

\bibitem{Dang2021}
Z.~Dang, C.~Deng, X.~Yang, K.~Wei, and H.~Huang, ``Nearest neighbor matching for deep clustering,'' in \emph{Proceedings of the IEEE/CVF Conference on Computer Vision and Pattern Recognition (CVPR)}, June 2021, pp. 13\,693--13\,702.

\bibitem{Dong2022}
H.~Dong, W.~Ma, L.~Jiao, F.~Liu, and L.~Li, ``A multiscale self-attention deep clustering for change detection in sar images,'' \emph{IEEE Transactions on Geoscience and Remote Sensing}, vol.~60, pp. 1--16, 2022.

\bibitem{Barlow2021}
\BIBentryALTinterwordspacing
J.~Zbontar, L.~Jing, I.~Misra, Y.~LeCun, and S.~Deny, ``Barlow twins: Self-supervised learning via redundancy reduction,'' in \emph{Proceedings of the 38th International Conference on Machine Learning}, ser. Proceedings of Machine Learning Research, M.~Meila and T.~Zhang, Eds., vol. 139.\hskip 1em plus 0.5em minus 0.4em\relax PMLR, 18--24 Jul 2021, pp. 12\,310--12\,320. [Online]. Available: \url{https://proceedings.mlr.press/v139/zbontar21a.html}
\BIBentrySTDinterwordspacing

\bibitem{Lv2021}
J.~Lv, Z.~Kang, X.~Lu, and Z.~Xu, ``Pseudo-supervised deep subspace clustering,'' \emph{IEEE Transactions on Image Processing}, vol.~30, pp. 5252--5263, 2021.

\bibitem{Niu2022}
C.~Niu, H.~Shan, and G.~Wang, ``Spice: Semantic pseudo-labeling for image clustering,'' \emph{IEEE Transactions on Image Processing}, vol.~31, pp. 7264--7278, 2022.

\bibitem{Zhang2021}
X.~Zhang, Y.~Ge, Y.~Qiao, and H.~Li, ``Refining pseudo labels with clustering consensus over generations for unsupervised object re-identification,'' in \emph{Proceedings of the IEEE/CVF Conference on Computer Vision and Pattern Recognition (CVPR)}, June 2021, pp. 3436--3445.

\bibitem{Wang2021}
L.~Wang, D.~Guo, G.~Wang, and S.~Zhang, ``Annotation-efficient learning for medical image segmentation based on noisy pseudo labels and adversarial learning,'' \emph{IEEE Transactions on Medical Imaging}, vol.~40, no.~10, pp. 2795--2807, 2021.

\bibitem{Karimireddy2020}
S.~P. Karimireddy, S.~Kale, M.~Mohri, S.~J. Reddi, S.~U. Stich, and A.~T. Suresh, ``Scaffold: Stochastic controlled averaging for federated learning,'' in \emph{Proceedings of the 37th International Conference on Machine Learning}, ser. ICML'20.\hskip 1em plus 0.5em minus 0.4em\relax JMLR.org, 2020.

\bibitem{Durmus2021}
\BIBentryALTinterwordspacing
D.~A.~E. Acar, Y.~Zhao, R.~Matas, M.~Mattina, P.~Whatmough, and V.~Saligrama, ``Federated learning based on dynamic regularization,'' in \emph{International Conference on Learning Representations}, 2021. [Online]. Available: \url{https://openreview.net/forum?id=B7v4QMR6Z9w}
\BIBentrySTDinterwordspacing

\bibitem{chen2023best}
H.~Chen, H.~Vikalo \emph{et~al.}, ``The best of both worlds: Accurate global and personalized models through federated learning with data-free hyper-knowledge distillation,'' \emph{arXiv preprint arXiv:2301.08968}, 2023.

\bibitem{Yao2010}
B.~Yao and L.~Fei-Fei, ``Grouplet: A structured image representation for recognizing human and object interactions,'' in \emph{2010 IEEE Computer Society Conference on Computer Vision and Pattern Recognition}, 2010, pp. 9--16.

\bibitem{VOC2012}
M.~Everingham, L.~Van~Gool, C.~K.~I. Williams, J.~Winn, and A.~Zisserman, ``The {PASCAL} {V}isual {O}bject {C}lasses {C}hallenge 2012 {(VOC2012)} {R}esults,'' http://www.pascal-network.org/challenges/VOC/voc2012/workshop/index.html.

\end{thebibliography}
% Generated by IEEEtran.bst, version: 1.14 (2015/08/26)

\vfill

\end{document}